\documentclass[conference]{IEEEtran}
\usepackage{times}

\usepackage[numbers]{natbib}
\usepackage{amsmath,mathtools}
\usepackage{amsfonts}
\usepackage{multicol}
\usepackage{xspace}
\usepackage[bookmarks=true]{hyperref}
\usepackage{color, colortbl}
\usepackage[dvipsnames]{xcolor}
\usepackage{graphicx}
\usepackage{caption}
\usepackage{cuted}
\usepackage{booktabs} 
\usepackage{multirow} 
\usepackage{wrapfig}
\usepackage{listings}
\usepackage{xcolor}
\usepackage{booktabs}

\newcommand{\aria}{Project Aria glasses\xspace}
\newcommand{\egoverse}{EgoVerse\xspace}
\newcommand{\egoversea}{\texttt{EgoVerse-A}\xspace}
\newcommand{\egoversei}{\texttt{EgoVerse-I}\xspace}
\newcommand{\egoversedb}{{EgoDB}\xspace}

\newcommand{\robotGT}{\texttt{Robot A}\xspace}
\newcommand{\robotStanford}{\texttt{Robot B}\xspace}
\newcommand{\robotUCSD}{\texttt{Robot C}\xspace}

\definecolor{reviewerAColor}{RGB}{180,30,30}
\definecolor{reviewerBColor}{RGB}{30,90,180}
\definecolor{reviewerCColor}{RGB}{0,140,90}
\definecolor{reviewerDColor}{RGB}{160,100,20}

\begin{document}

\title{\egoverse: An Egocentric Human Dataset for \\Robot Learning from Around the World
}


\author{
Ryan Punamiya$^{*1}$, Simar Kareer$^{*1}$,
Zeyi Liu$^{2}$, Josh Citron$^{2}$,
Ri-Zhao Qiu$^{3}$, Xiongyi Cai$^{3}$,
Alexey Gavryushin$^{4}$, \\ 
Jiaqi Chen$^{4}$, Davide Liconti$^{4}$ 
Lawrence Y. Zhu$^{1}$, Patcharapong Aphiwetsa$^{1}$, Baoyu Li$^{1}$,
Aniketh Cheluva$^{1}$, \\
Pranav Kuppili$^{1}$, Yangcen Liu$^{1}$,
Dhruv Patel$^{1}$, Aidan Gao$^{1}$, Hye-Young Chung$^{1}$, Ryan Co$^{1}$,
Renee Zbizika$^{2}$,\\
 Jeff Liu$^{2}$, Xiaomeng Xu$^{2}$, Haoyu Xiong$^{5}$,
Geng Chen$^{3}$,
Sebastiano Oliani$^{4}$, Wenkai Xuan$^{4}$, Chenyu Yang$^{4}$, Xi Wang$^{4}$,\\
James Fort$^{6}$, Richard Newcombe$^{6}$,
Josh Gao$^{7}$, Jason Chong$^{7}$,
Garrett Matsuda$^{8}$, Aseem Doriwala$^{8}$,\\
Marc Pollefeys$^{4}$, Robert Katzschmann$^{4}$,
Xiaolong Wang$^{3}$, Shuran Song$^{2}$,
Judy Hoffman$^{1}$, Danfei Xu$^{1}$ \\
\\
$^{1}$Georgia Institute of Technology,
$^{2}$Stanford University, 
$^{3}$University of California San Diego,
$^{4}$ETH Zürich \\
$^{5}$MIT CSAIL,
$^{6}$Meta Reality Labs Research,
$^{7}$Mecka AI,
$^{8}$Scale AI

}


%

\maketitle
\begin{strip}
\vspace{-3em}
\centering
\includegraphics[width=\textwidth]{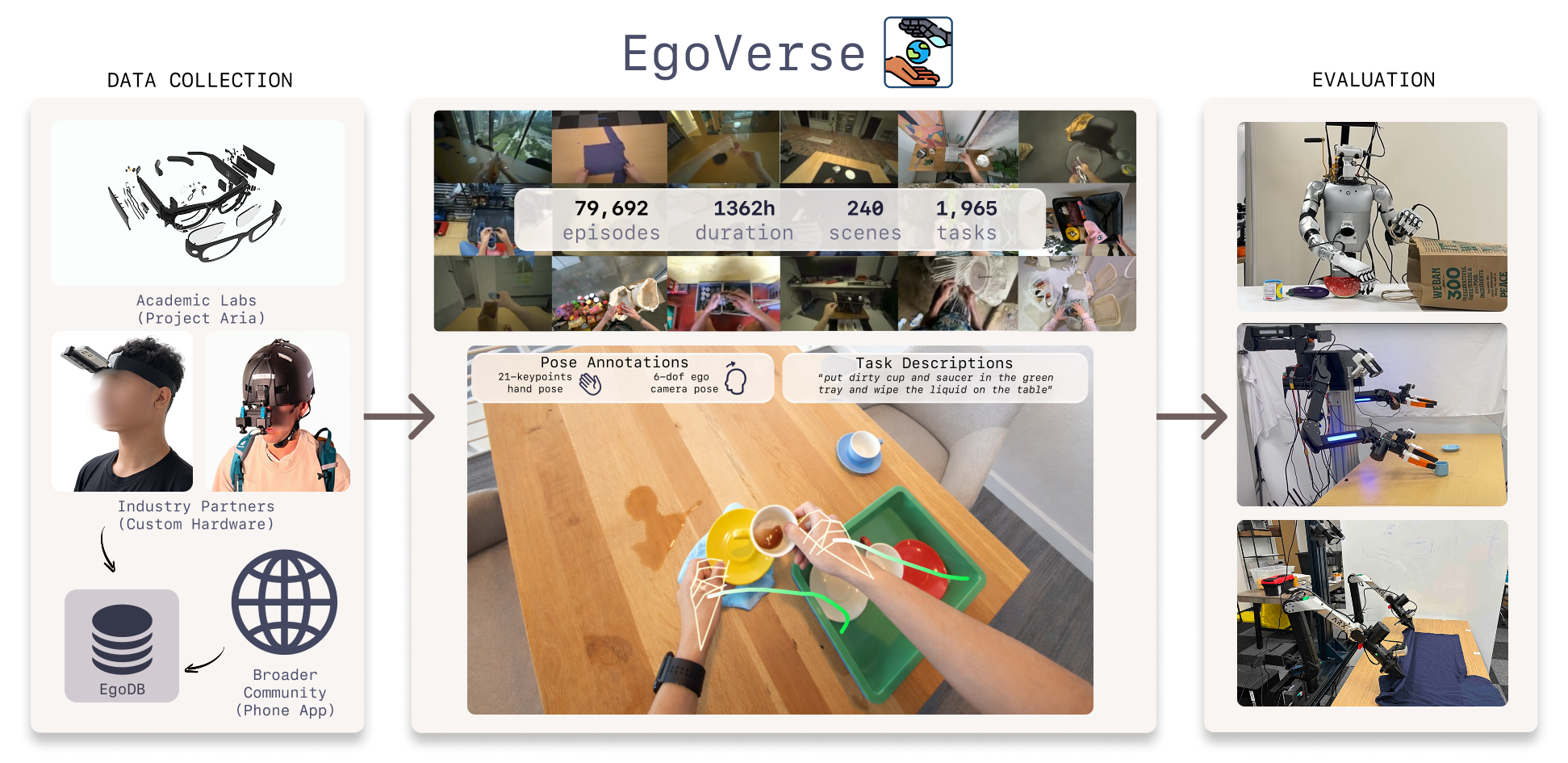}
\captionof{figure}{\textbf{Overview.} \egoverse is a collaborative framework for scalable human data–driven robot learning.
\textbf{Capture:} Egocentric demonstrations are collected worldwide using academic, industry, and community-accessible hardware systems, continuously aggregated by a centrally-hosted data management system.
\textbf{Dataset:} All data are unified into a shared dataset with egocentric video, 3D hand poses, camera motion, and task descriptions across diverse tasks and scenes.
\textbf{Evaluation:} This work presents a large-scale evaluation study on human-to-robot transfer with shared protocols across multiple labs and robot embodiments.}
\vspace{-1em}
\label{fig:teaser}

\end{strip}

\begin{abstract}
Robot learning increasingly depends on large and diverse data, yet robot data collection remains expensive and difficult to scale. Egocentric human data offer a promising alternative by capturing rich manipulation behavior across everyday environments. However, existing human datasets are often limited in scope, difficult to extend, and fragmented across institutions.
We introduce EgoVerse, a collaborative platform for human data–driven robot learning that unifies data collection, processing, and access under a shared framework, enabling contributions from individual researchers, academic labs, and industry partners. The current release includes 1,362 hours (80k episodes) of human demonstrations spanning 1,965 tasks, 240 scenes, and 2,087 unique demonstrators, with standardized formats, manipulation-relevant annotations, and tooling for downstream learning.
Beyond the dataset, we conduct a large-scale study of human-to-robot transfer with experiments replicated across multiple labs, tasks, and robot embodiments under shared protocols. We find that policy performance generally improves with increased human data, but that effective scaling depends on alignment between human data and robot learning objectives. Together, the dataset, platform, and study establish a foundation for reproducible progress in human data–driven robot learning.
\end{abstract}

\IEEEpeerreviewmaketitle

\vspace{-1em}
\section{Introduction}

Recent progress in robot learning has shown that scaling data is a powerful driver of generalization \cite{black2410pi0, hu2024data, khazatsky2024droid, o2024open}. Large-scale imitation learning has enabled policies to handle broader task distributions, more visual variation, and longer horizons, echoing trends seen in large vision and language models. However, unlike those domains, robot learning faces a fundamental bottleneck -- collecting robot demonstrations requires physical hardware, expert teleoperation, and controlled setups. As a result, expanding robot datasets in scale and diversity remains slow, expensive, and difficult to sustain.

In contrast, egocentric human data offers a promising alternative. Humans naturally perform manipulation and loco-manipulation tasks across diverse environments on a daily basis, generating rich behavioral data at a scale that is infeasible for robots alone. Importantly, human data also provides a unifying abstraction for the community. Instead of coordinating around a specific robot embodiment, researchers can focus on curating diverse, real-world experience data while deferring embodiment decisions downstream. This property has driven growing academic and commercial interest in leveraging egocentric human demonstrations, supported by recent advances in wearable sensors \cite{engel2023projectarianewtool,egodex,yin2025osmo} and large-scale data capture systems~\cite{chi2024universal,etukuru2025robot}.

Despite the promise of leveraging human data, two major challenges remain. First, effective human-robot transfer remains an open research problem, with unresolved questions around the embodiment gap and scaling behavior. Second, most existing human datasets are one-off, static releases collected for a specific study, making further scaling difficult \cite{egodex, kareer2024egomimicscalingimitationlearning, qiu2025-humanpolicy}. Addressing these limitations requires more than collecting a larger dataset: it calls for a continuously growing human data ecosystem that can evolve with new contributors and provide durable insights into human-to-robot transfer.

This paper introduces \textbf{EgoVerse}, a large-scale collaborative framework to create an ever-growing dataset of egocentric human demonstrations specifically designed for robot learning, paired with a systematic study of the key factors that enable effective cross-embodiment transfer from diverse human data sources. Our contributions are threefold:

\textbf{The \egoverse Dataset.} We present a large dataset of egocentric human demonstrations contributed by a consortium of academic groups and industry partners worldwide. The dataset is intentionally composed of \textbf{two complementary parts} with distinct purposes. The first, named \egoversea, consists of data collected under carefully controlled and standardized protocols, mirrored across participating academic labs. This component is designed to enable \emph{reproducible studies} and \emph{systematic analysis}. The second, named \egoversei, focuses on scale, diversity, and richness of annotation, and is sourced from industry partners collecting data in the wild. This industry-driven stream enables EgoVerse to grow beyond the limits of academic institutions and to align research with real-world deployment and industrial interests. Overall, EgoVerse contains 1,362 hours of human demonstrations across 240 scenes, and 1965 tasks, and 2087 demonstrators. Across both components, the dataset captures manipulation tasks with unprecedented diversity across scenes, objects, and human demonstrators, while maintaining consistency through shared task semantics, standardized protocols, and high-quality manipulation-relevant signals, including 3D hand and head poses, and subtask-level descriptions.

\textbf{The \egoverse Ecosystem.} \egoverse is designed to grow over time with contributions from both academic labs and industry partners. To unify this evolving collection, we introduce \textbf{\egoversedb}, a scalable data management and access system that supports continuous data ingestion from diverse sources, including individual contributors, academic labs, and industry partners. Unlike prior static datasets~\cite{hot3d, fan2023arctic, egodex, hoi4d}, \egoversedb enables the notion of a \emph{living dataset} that grows and evolves with ongoing contributions. The system provides standardized data processing, unified storage formats, controlled data access, visualization tools, and interfaces for downstream learning algorithms.  
In addition to lab- and partner-operated capture systems, \egoverse includes a \textbf{phone-based human data collection pipeline} that enables lightweight egocentric recording using commodity smartphones. 
Together, the \egoverse ecosystem lowers the barrier for groups and individuals with limited resources while also enabling contributors to share data back with the community in a structured and reproducible way.

\textbf{A Consortium-Scale Study.} We conduct the most comprehensive study to date on what matters when learning robot manipulation policies from diverse human embodiment data. Our study is \textbf{reproducible by design}: experiments are intentionally replicated across multiple independent labs, tasks, and robot embodiments using shared protocols and evaluation criteria. In particular, the results are executed on \textbf{three distinctive robot embodiments} to ensure the main findings are not system-specific. This cross-lab, cross-embodiment setup allows us to identify conclusions that are consistent and robust across settings, as well as effects that systematically differ due to embodiment, sensing, or control variations. 

\textbf{Key Findings}. 
Our study yields several consistent findings across labs, tasks, and robot embodiments. First, co-training robot policies with human data leads to clear and reproducible performance improvements. To our knowledge, this is the first time this effect is validated under a standardized, cross-lab experimental setup spanning multiple robots. Second, the benefits of scaling human data depend critically on the availability of aligned human–robot data, where human and robot data share task semantics and scene context. We find that positive scaling emerges only when such aligned data are included as part of training, suggesting it provides essential grounding for effective human-to-robot transfer. Finally, we find that different forms of human data diversity contribute unevenly to generalization. Increasing demonstrator diversity improves robustness to unseen human embodiments, while scene diversity plays a dominant role in generalization to novel environments, particularly under limited data budgets.

\section{Related Work}
\noindent\textbf{Datasets of Human Activities.}
Large-scale human activity datasets such as Something-Something V2~\cite{sth-sth}, Ego4D~\cite{grauman2022ego4d}, HOI4D~\cite{hoi4d}, EgoExo4D~\cite{EgoExo4D}, and Epic-Kitchens~\cite{Epic-Kitchens} capture rich human behavior across diverse environments. However, they are not designed for robot learning. These datasets often include tasks beyond current robot capabilities, lack manipulation-relevant annotations such as precise hand poses or object interactions, and contain unstructured activities that are difficult to translate into executable robot demonstrations. In contrast, our approach emphasizes ``bounded diversity'', focusing on tasks that are feasible for typical bimanual mobile manipulators while preserving natural variation across environments, objects, and demonstrators. More recent datasets~\cite{hot3d,egodex} introduce manipulation-relevant sensing, including camera calibration and hand pose tracking. While these represent important progress, they are typically released as static, study-specific datasets collected over limited time spans and environments, making them difficult to extend. Moreover, these works do not include systematic evaluation of transfer to robot learning. \egoverse takes a different approach by treating human data as a continuously growing research resource, and by explicitly validating the dataset through a large-scale, reproducible study designed to ensure robot-learning readiness.

\noindent\textbf{Robot Learning from Human Data.} Human data presents two main opportunities for robot learning: abundant unlabeled online videos and curated, labeled demonstrations~\cite{bahl2022humantorobotimitationwild, kareer2024egomimicscalingimitationlearning, avid, lbw}. Web videos, though plentiful, require pseudo-labeling of actions via inverse dynamics models~\cite{bu2025univla, UniPi, ye2024latentactionpretrainingvideos}, affordances~\cite{bahl2023affordances, shi2025zeromimic}, or point tracking~\cite{bharadhwaj2024track2act, ren2025motiontracksunifiedrepresentation, wen2023anypoint} for policy training, forming a basis for some foundation models~\cite{chen2025largevideoplanner, beingbeyond2025beingh0, nvidia2025gr00tn1openfoundation, yang2025egovlalearningvisionlanguageactionmodels}, yet often still necessitating in-domain robot data. Alternatively, labeled human demonstrations can be co-trained with robot data as distinct embodiments for policy learning~\cite{guzey2025aina, kareer2024egomimicscalingimitationlearning, lepert2025phantomtrainingrobotsrobots, immimic, qiu2025-humanpolicy, zhu2026emma}, post-training~\cite {cai2025n, kareer2025emergencehumanrobottransfer}, and world modeling~\cite{goswami2025dexwm, he2025scalingcrossembodimentworldmodels}.  These works found that this practice enhances robustness and scene understanding. However, such findings remain confined to limited scale and single robot embodiment, leaving critical questions about multiple robot embodiments and varied human data sources largely unexplored. Our work addresses these fundamental gaps through a large-scale human dataset and a carefully-executed consortium-scale study.

\noindent\textbf{Scaling Robot Learning with Massive Data.} Recent progress in foundational policy models highlights the benefits of scaling with large datasets. Public efforts such as Open X-Embodiment~\cite{o2024open}, DROID~\cite{khazatsky2024droid}, and Rh20t~\cite{fang2023rh20t} demonstrate that training on diverse, multi-embodiment data improves generalization across tasks and environments. However, achieving generally capable robots remains fundamentally constrained by data scalability, as robot teleoperation is expensive and labor-intensive.
Our work instead examines how human egocentric data can support robot learning at scale and study it as a \emph{first-class data source} alongside robot data.

\section{The EgoVerse Dataset: A Human Dataset for Robot Learning from Around the World}
\label{sec:dataset}

\subsection{Human Data Collection Setup}
\label{ssec:dataset:data_hardware}

Academic partners used \aria as the standard device for \egoversea. Industry partners contributed to \egoversei with custom-built rigs for scalability and ease of deployment. We also introduce a phone-based capture system that is accessible by the broader community. The setups are illustrated in Fig.~\ref{fig:aria_setup}, with more detail in Appendix.

\noindent \textbf{\aria for human data collection.}
Following prior work~\cite{kareer2024egomimicscalingimitationlearning, egozero, punamiyaegobridge, zhu2026emma}, we adopt \aria (Gen~1) as the standardized capture platform for \egoversea. \aria are lightweight (75 g) head-worn devices with a wide-FoV RGB camera and two synchronized monochrome scene cameras used for SLAM and hand tracking (Fig.~\ref{fig:aria_setup}). The side cameras maintain visibility of hand motion even when out of the forward-looking RGB view.

\noindent\textbf{Industry partner data collection setup.}
The \egoversei stream aggregates demonstrations collected using custom wearable sensor platforms deployed in diverse indoor environments. These rigs typically feature lightweight head-mounted cameras paired with synchronized sensing modules. Most systems use stereo fisheye RGB cameras to achieve accurate hand pose estimation. The data include ego-view RGB videos and, where available, depth streams, with additional inertial sensing to reconstruct camera motion. 

\noindent\textbf{Phone-based capture system.} To broaden access to data capture devices, we also make available a setup based on commodity smartphones as part of the \egoverse ecosystem. This system uses an iPhone mounted on a head strap, with the ultrawide camera recording egocentric RGB video at 1080p and 30 FPS. Captured videos are uploaded to a cloud-based processing pipeline that recovers 6-DoF head pose via visual tracking and estimates 3D hand poses with 21 keypoints per hand. More details on the system are in the Appendix.

\begin{figure}[t]
\centering
\includegraphics[width=\linewidth]{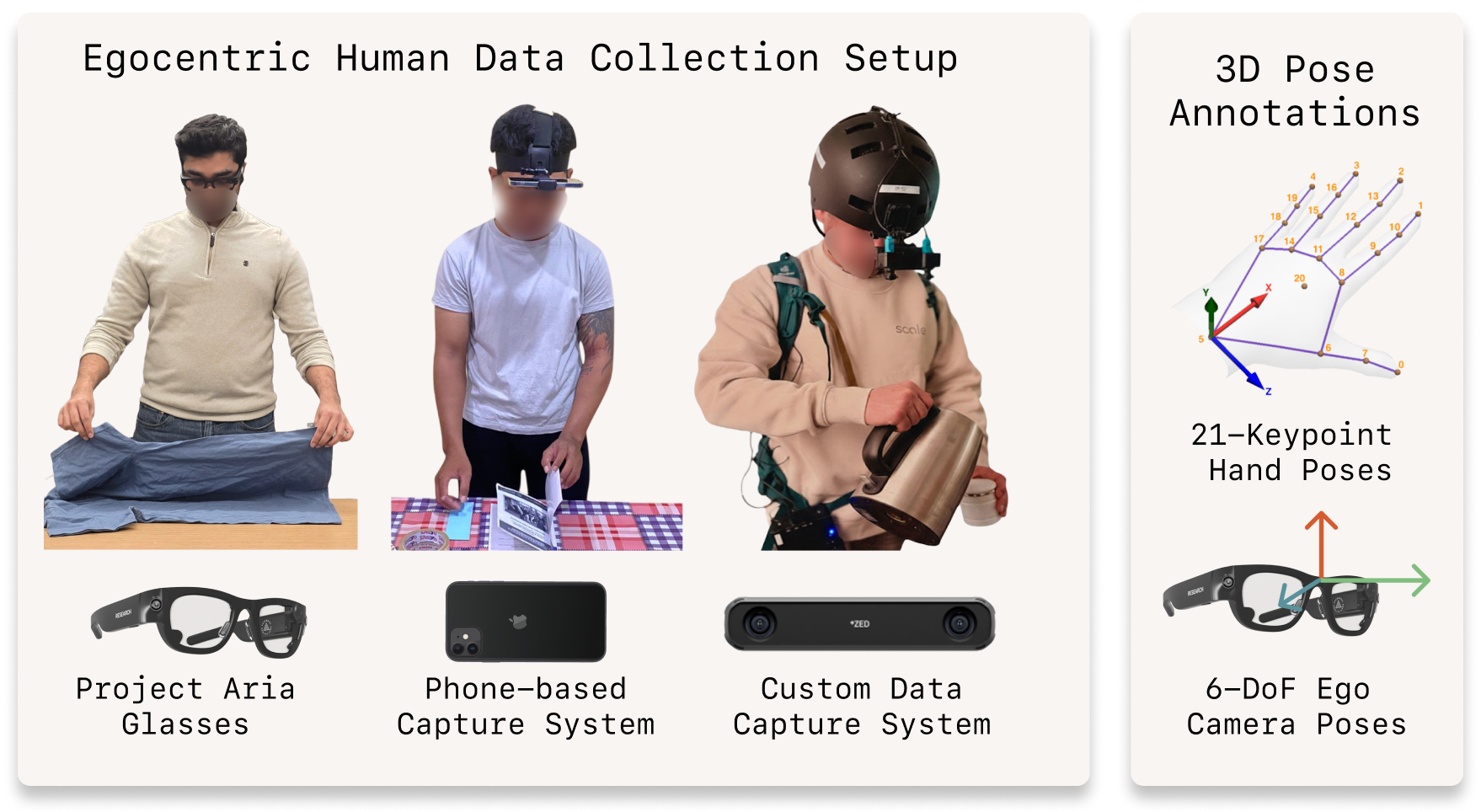}
\caption{\textbf{Human Data Capture Setup.} (Left) \egoverse is captured through a variety of hardware systems, including \aria (academic labs), a phone-based capture system (accessible by everyone), and custom setups by industry partners. (Right) Regardless of sources, human data is processed into a unified format that contains at minimum egocentric videos, hand keypoints, and camera poses.}
\label{fig:aria_setup}
\vspace{-10pt}
\end{figure}

\subsection{Human Data Annotation}
\label{ssec:dataset:processing}

EgoVerse augments egocentric human demonstrations with structured annotations tailored for robot policy learning. For each frame, we estimate 3D hand pose for both hands using 21 keypoints per hand in the camera frame, paired with a calibrated 6-DoF head pose obtained from visual–inertial SLAM. Academic partners use Project Aria’s Machine Perception Service (MPS) for tracking and egomotion, while industry datasets combine partner SLAM, model-based pose estimation, and post-processing to ensure consistent trajectories. In our human-to-robot transfer studies (Sec.~\ref{sec:study}), these signals serve as proxies for human end-effector motion, enabling cotraining across embodiments. Beyond poses, annotation granularity differs across dataset components. \egoversea follows a lightweight per-episode protocol, attaching task descriptions, scene identifiers, primary manipulated objects, and demonstrator metadata to support controlled cross-lab analysis. In contrast, \egoversei provides denser annotations, including fine-grained (1–2 s) language descriptions, active-hand indicators, static versus mobile manipulation flags, and additional contextual tags where available.

\subsection{\egoversedb: Scalable Data Management System}
\begin{figure}[t]
\centering
\includegraphics[width=0.75\linewidth]{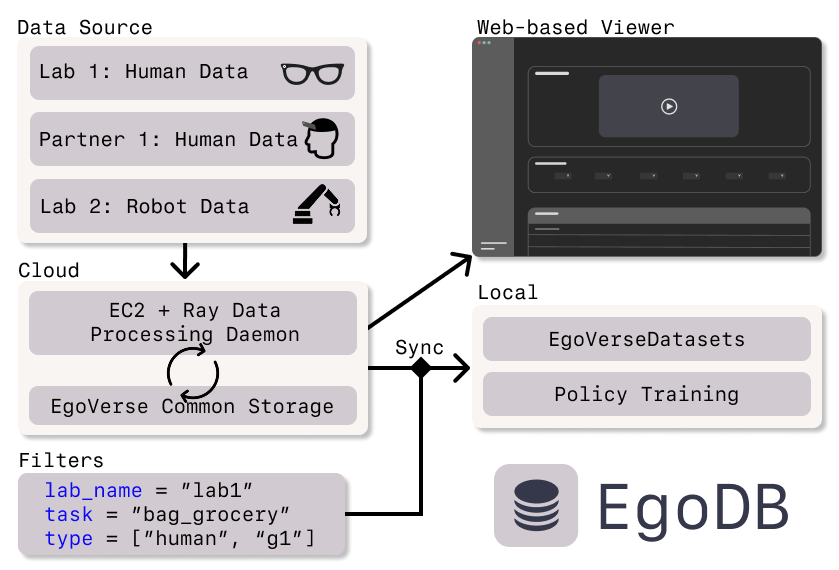}
\caption{\textbf{\egoversedb}. Human and robot data from multiple labs and partners are ingested into a cloud-based processing pipeline, unified in a common storage format, and made accessible through a web-based viewer. Users can sync filtered subsets of the dataset to local machines for downstream policy training.}
\label{fig:egoversedb}
\vspace{-10pt}
\end{figure}

To support consortium-wide collaboration and long-term dataset growth, we develop \egoversedb, a cloud-based system for continuous ingestion, processing, and access of heterogeneous human and robot data (Fig.~\ref{fig:egoversedb}). Data from distributed sources are uploaded to S3-backed storage and converted into a unified, training-ready format shared across the \egoverse ecosystem. A nightly pipeline performs standardized preprocessing, validation, and indexing to ensure consistent usability for downstream learning. Episode metadata are registered in a centralized SQL database, enabling structured queries over tasks, embodiments, scenes, data sources, and annotation types. \egoversedb also provides a web-based interface for browsing demonstrations, inspecting annotations, and tracking dataset growth. For local training, users can synchronize filtered subsets of the dataset via configuration files, enabling reproducible access without manual data management.

\subsection{The \egoversea Dataset}
\label{ssec:dataset:protocol}
\noindent\textbf{Collection Protocol and Dataset Units.}
To ensure standardized data collection across geographically distributed sites, we adopt a shared protocol organized around \emph{dataset units}. Each unit follows a common instruction format to ensure consistent task execution across labs, and typically consists of approximately 5 minutes of recording, yielding 5–10 demonstrations per task. We enforce basic quality constraints to maintain visibility of hands and manipulated objects, and log key metadata such as demonstrator identity, scene, and object set for traceability. Additional protocol details and quality control procedures are provided in the appendix.

\begin{figure*}[t]
\centering
\includegraphics[width=\textwidth]{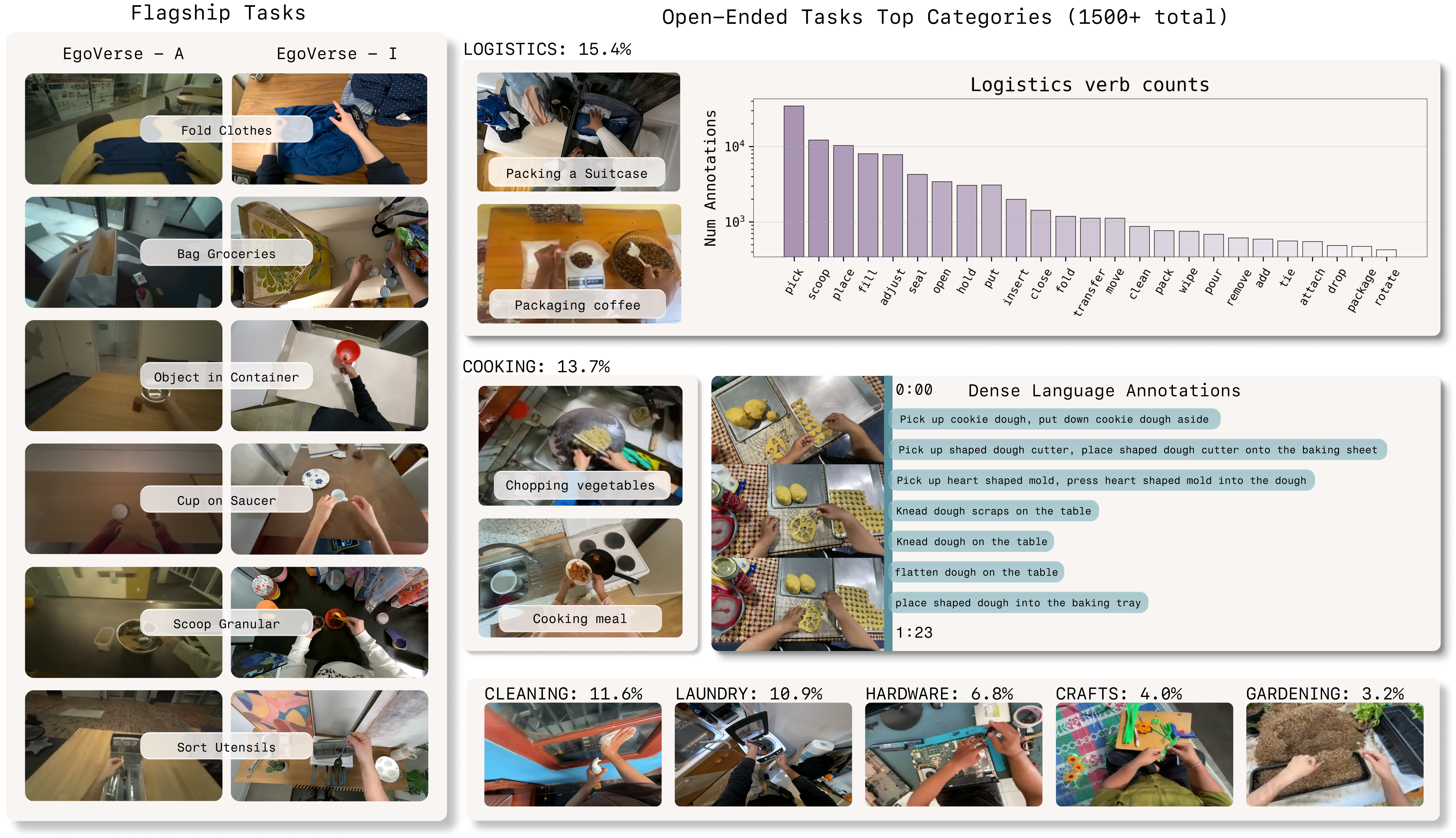}
\caption{\textbf{Dataset Composition and Diversity.}
Left: \egoversea and \egoversei include six shared flagship manipulation tasks collected across diverse scenes and demonstrators. Right: \egoversei contains over 1,500 open-ended tasks spanning everyday activity categories, with representative verb frequency distributions illustrating the diversity of manipulation actions.
}
\label{fig:egoverse_vis}
\vspace{-10pt}

\end{figure*}

\noindent\textbf{Flagship Tasks.}
To control task semantics while allowing other axes of variation, we define a set of six \emph{Flagship Tasks} shared across all participating labs. 

\begin{itemize}
\item \textit{object-in-container}: Pick, place into a container, dump, and repeat continuously for 40 seconds with randomized object--container pairs. Single-arm task.
\item \textit{cup-on-saucer}: Reorient a cup from a random initial pose and place it on a saucer. Bimanual task.
\item \textit{bag-grocery}: Open a grocery bag and load 1--3 items. Bimanual task.
\item \textit{fold-clothes}: Three-fold a T--shirt initialized in random configurations. Bimanual task.
\item \textit{scoop-granular}: Scoop granular material (e.g., beans) and transfer it to a container until full. Single-arm task.
\item \textit{sort-utensils}: Pick and sort utensils into designated containers. Single-arm task.
\end{itemize}

These tasks span a range of manipulation regimes, including single-arm and bimanual coordination, fine-grained object placement, and longer-horizon behaviors, while remaining feasible for common robot manipulation platforms.

\noindent\textbf{Structured Axes of Diversity.}
\egoversea is designed to capture diversity while maintaining controlled task semantics and data quality. Human data collection is organized along three primary axes: \emph{task}, \emph{scenario}, and \emph{demonstrator}.

\emph{Scenario and object diversity.}
Each flagship task is performed across 8–12 scenes per lab, with 1–10 dataset units collected per scene to capture within-environment variation. Demonstrations are recorded within a roughly $40\text{cm} \times 60\text{cm}$ workspace, with object positions randomized across trials. Objects are sampled from a fixed set of up to 30 per task within each lab, while independent procurement across sites introduces substantial variation in object geometry, appearance, and material properties.

\emph{Demonstrator diversity.}
Data are collected from 1–8 demonstrators per lab. Despite identical instructions, demonstrators within and across labs exhibit consistent differences in motion patterns, timing, coordination strategies, and hand trajectories. This naturally induces variation in human morphology and egocentric viewpoints due to differences in height, posture, and workspace configuration. Prior work shows that such human variability can significantly influence policy learning and cross-embodiment alignment~\cite{kareer2024egomimicscalingimitationlearning, immimic, punamiyaegobridge}. Rather than eliminating this variation, we treat it as an inherent property of scalable human data.

\noindent\textbf{Controlled-Diversity Subset.}
While cross-lab aggregation provides realistic diversity, it also introduces uneven coverage across scenes and demonstrators. To study the effects of diversity under controlled conditions, one lab collected a secondary dataset for \textit{cup-on-saucer} and \textit{fold-clothes} using a fixed pool of 16 demonstrators and 16 scenes. Data were allocated via a structured assignment matrix (3.75 minutes to 2 hours per demonstrator--scene pair; $\ge$1 hour per scene) across three experimental regimes:


\begin{itemize}
    \item \textbf{Single-Scene Demonstrator Scaling.} 
    A fixed data budget of 2 hours is varied in a single scene as the number of demonstrators increases from 1 to 16 to study the impact of demonstrator diversity in a single scene.

    \item \textbf{Multi-Scene Interaction Effects.} Scenes 1--8 are used to study demonstrator scaling across diverse environments, where the number of demonstrators is varied from 4--12 with a fixed data budget of 8 hours.
    
    \item \textbf{Scene Diversity Scaling.}
    Scenes 1--16 are collected in a fixed demonstrator pool, decoupling scene diversity and data quantity.
\end{itemize}

By enforcing control over sources of variation, this subset allows scene diversity and demonstrator diversity to be independently scaled and studied (Sec.~\ref{sec:experiment:diversity}).

\subsection{The \egoversei Dataset}
\label{ssec:dataset:industry_data}
\vspace{-1em}

\egoversea densely covers the flagship tasks along controlled axes but lacks the task diversity required to train generalist policies. To address this, we introduce \egoversei, the largest action-labeled egocentric human dataset, comprising nearly 1,400 hours of data across nearly 2,000 tasks, 240 scenes, and 2,087 demonstrators (Fig.~\ref{fig:egoverse_vis}). \egoversei is designed to be 
\begin{wrapfigure}{l}{0.245\textwidth}
\includegraphics[width=\linewidth]{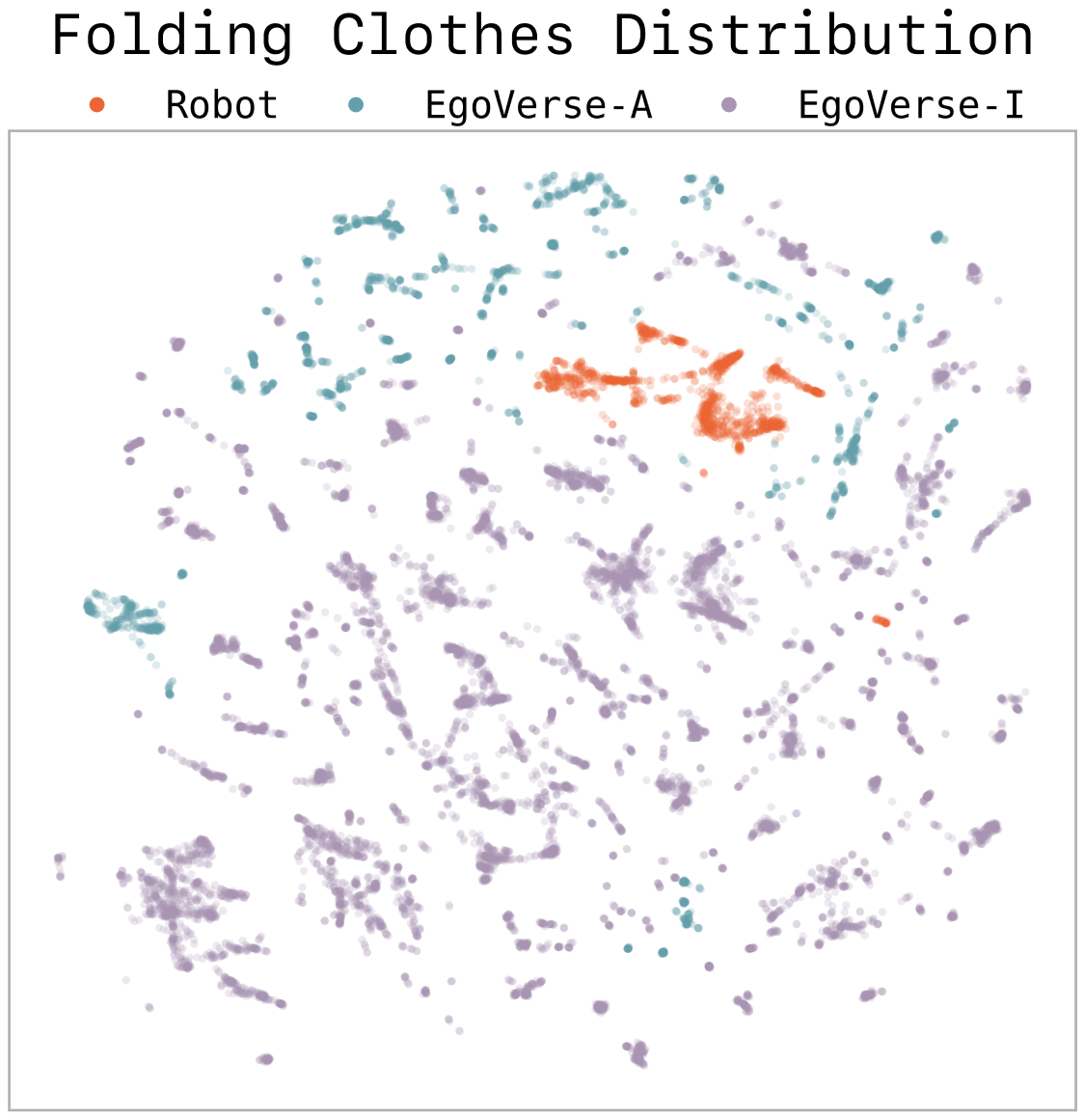}
\caption{UMAP of DINOv3 embeddings.}
\label{fig:umap_diversity}
\vspace{-1em}
\end{wrapfigure}
maximally useful for robot learning by emphasizing manipulation-heavy task distributions, instructing demonstrators to keep hands visible, and applying manual quality control to retain only manipulation-dense segments. 
In addition, \egoversei includes dense language annotations (Sec.~\ref{ssec:dataset:processing}), making it suitable for training language-conditioned policies such as VLAs. Fig.~\ref{fig:egoverse_vis} compares flagship task examples from \egoversea and \egoversei, and illustrates representative language annotations and verb statistics for the open-ended tasks in \egoversei.


\noindent\textbf{Visual Diversity Case Study.} To compare visual diversity across data sources, we apply UMAP to DINOv3 embeddings from the \emph{fold-clothes} task across three sources: robot data from a single lab, human data in \egoversea, and human data in \egoversei. As shown in Fig.~\ref{fig:umap_diversity}, despite shared task semantics, \egoversei substantially expands the visual coverage. This increased diversity is critical for learning policies that generalize beyond lab-specific appearance statistics.






\section{The EgoVerse Study: A Consortium-Scale Examination of Human-to-Robot Transfer}
\label{sec:study}
We conduct a large-scale systematic study to examine the factors that enable effective transfer from diverse human embodiment data to robot manipulation policies. Our experimental design prioritizes \emph{reproducibility}: robot experiments are replicated across multiple labs using distinct platforms, controllers, and environments. This multi-lab, multi-embodiment setup enables identification of findings that are consistent across hardware and sensing configurations, ensuring that our conclusions extend beyond system-specific effects. Using shared protocols and evaluation criteria, we study how human data scale and diversity influence human-to-robot transfer.

\subsection{Robot Platforms}
\label{ssec:dataset:robots}

\begin{figure}[t]
\centering
\includegraphics[width=\linewidth]{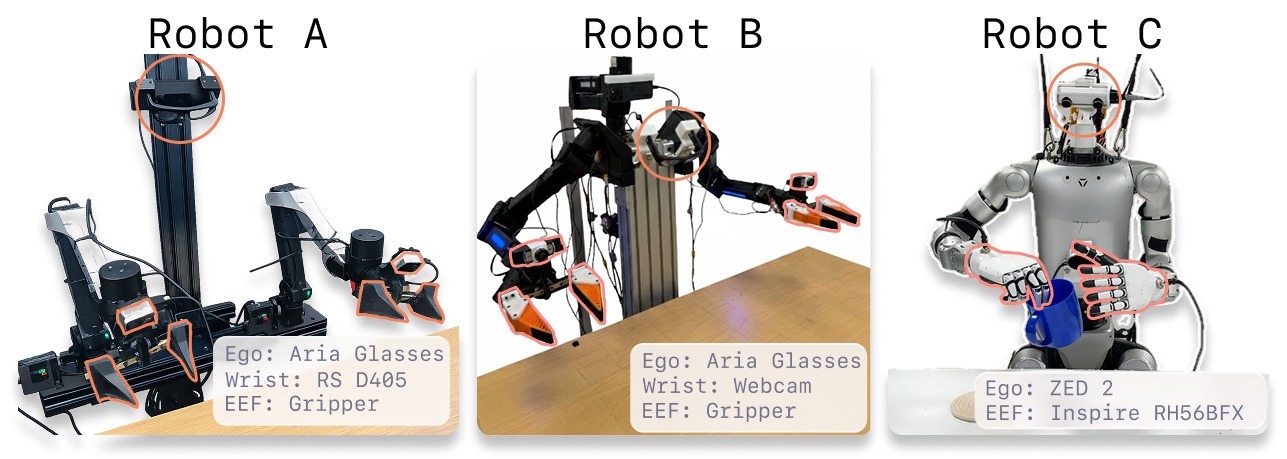}
\caption{\textbf{Robot Platforms.} We perform evaluation on three distinctive robot platforms across labs with shared protocols.}
\label{fig:robots}
\vspace{-10pt}
\end{figure}

Our experiments are carried out on three distinctive robot platforms with varied kinematics, sensing configuration and control interfaces. This allows us to assess which findings are consistent across systems rather than specific to a single robot. (Fig.~\ref{fig:robots})

{\robotGT.} The system comprises of two 6-DoF ARX5 robot arms with parallel jaw grippers. The arms are mounted upright. The main egocentric camera is an Aria Glasses, with two wrist-mounted Intel RealSense D405 cameras. 

{\robotStanford.} Different than Robot A, two ARX5 arms are side-mounted on a custom 3D-printed shoulder structure for human-like workspace~\cite{via}. The robot is equipped with a head-mounted Aria glass and wrist-mounted Logitech webcams on each end effector.

{\robotUCSD.} The system is a Unitree G1 robot with 7-DoF arms, each equipped with a 6-DoF Dexterous Inspire Hand. The main egocentric camera is a ZED 2 stereo camera.


\subsection{Human and Robot Data Alignment}

\noindent\textbf{Robot Actions Representations.} For \robotGT, base-frame $\mathbb{SE}(3)$ actions are computed from commanded joint angles via forward kinematics, projected into the egocentric camera frame using extrinsics, and represented as 6-DoF Euler poses ($x, y, z, \text{yaw, pitch, roll}$) with a gripper state for each arm, yielding $a^R_{t:t+k} \in \mathbb{R}^{k \times 14}$. \robotStanford follows a similar pipeline, but represents orientation using quaternions ($x, y, z, q_x, q_y, q_z, q_w$) along with the gripper state, resulting in $a^R_{t:t+k} \in \mathbb{R}^{k \times 16}$. For \robotUCSD, the wrist motion is modeled as an absolute $\mathbb{SE}(3)$ pose trajectory in the robot base frame, and dexterous hand control is specified via five keypoint positions relative to the end-effector, which are mapped to joint commands using an inverse kinematics solver.

\noindent\textbf{Human Action Representation. } Human egocentric hand tracking is usually with respect to a moving camera frame. Following prior work~\cite{kareer2024egomimicscalingimitationlearning,qiu2025-humanpolicy}, to unify the reference frames between human and robot for joint policy learning, we opt to construct camera-centered stable reference frames. The raw trajectory of $\mathbb{SE}(3)$ hand poses [$p_t^H$, $p_{t+1}^H$, $p_{t+2}^H$, $...$, $p_{t+k}^H$], where each pose $p_t^H$ is in the device frame $T_t^{device}$. We construct an action $a^H_{t:t+k}$ by projecting the future hand positions in to the $t$-th device frame. As such, the trajectory is constructed by 
\[
a^H_{t:t+k} = \left[ \left(T_t^{\text{device}}\right)^{-1} T_{t+i}^{\text{device}} \cdot p_{t+i}^H \right]_{i=1}^{k}
\]
\noindent\textbf{Aligning Human and Robot Data. } Recent cross-embodiment work~\cite{kareer2024egomimicscalingimitationlearning, punamiyaegobridge, octomodelteam2024octoopensourcegeneralistrobot, zhu2026emma} show that co-training benefits from individually normalizing proprioception and actions. To make the normalization robust to outliers, we employ \emph{quantile} normalization. We map the $1^{st}$ and $99^{th}$ percentiles of the feature distribution to the range $[-1, 1]$, following~\cite{chi2024diffusionpolicy, pi0.5}. For a feature tensor $x$, the normalized output $\hat{x}$ is calculated as 
$\hat{x} = 2 \cdot \left( \frac{x - q_{0.01}}{q_{0.99} - q_{0.01}} \right) - 1$. 
To account for varying camera sensors, we perform random image crop and color jittering during training.

\subsection{Learning Architecture and Algorithm}

\noindent\textbf{Policy Architecture.}
To enable joint training across diverse embodiments, we adopt an encoder–decoder architecture with shallow, modality-specific stems~\cite{wang2024hpt}. Image observations are processed by a ResNet-18~\cite{he2015deepresiduallearningimage} backbone, while proprioceptive inputs are encoded with an MLP, before being tokenized into a shared space via learned query attention. A shared vision stem processes egocentric RGB observations from both human and robot embodiments, while separate stems handle robot-specific wrist cameras and proprioception. The resulting tokens are concatenated and passed, together with a set of learnable tokens, through a shared transformer encoder $f_\phi$. These learnable tokens attend to the multi-modal inputs to extract task-relevant features and condition the action decoders.
\begin{figure}[h]
\centering
\includegraphics[width=\linewidth]{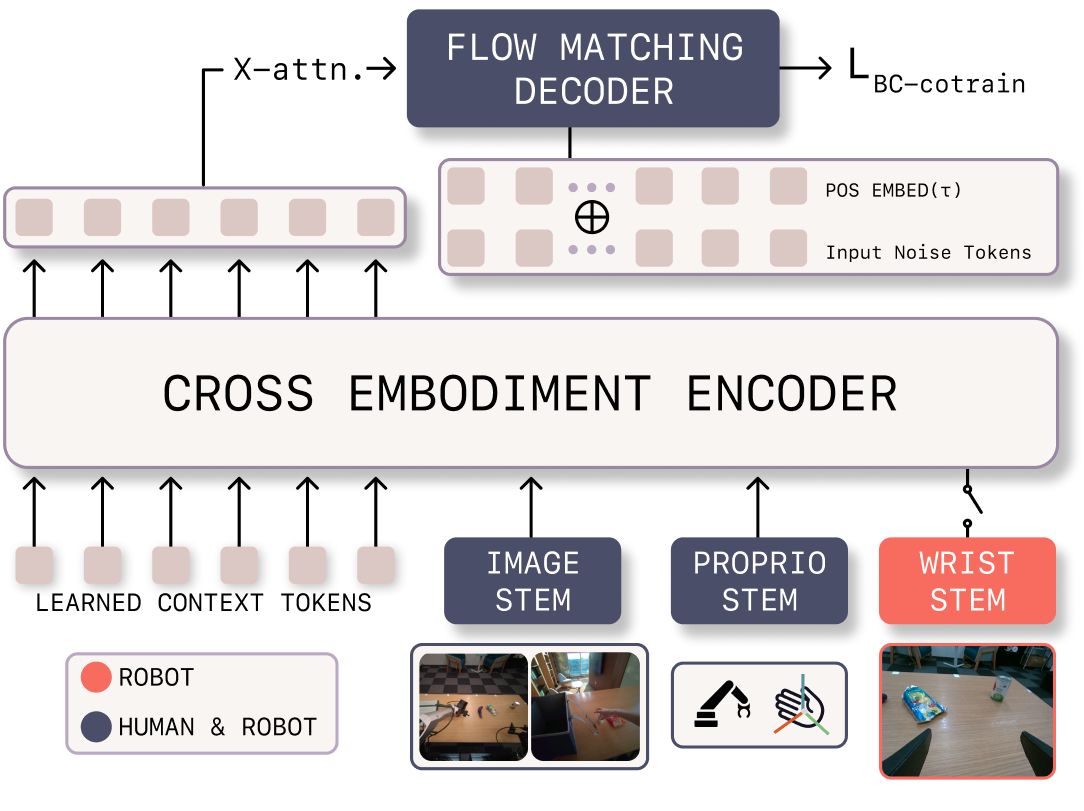}
\caption{\textbf{Model Architecture.} An illustration of our transformer-based cross-embodiment policy backbone.}
\label{fig:arch}
\end{figure}

\noindent\textbf{Flow Matching Action Decoder. } The action decoder $\pi_\theta$ is parameterized by a multi-block transformer decoder trained with a flow matching objective. $T$ learnable tokens which correspond to the action sequence are initialized. The timestep $\tau \sim \text{Beta}(1.5, 1.0)$ is  positionally embedded and concatenated to the learnable tokens along the hidden dimension. With alternating self and cross attention blocks, we inject the encoded context from the encoder. The learnable tokens are decoded into the action dimension using a linear layer. Depending on the output action space, we initialize shared or embodiment-specific action decoders. 

\noindent\textbf{Training Objective. } The encoder $f_\phi$ and the decoder $\pi_{\theta}$ are jointly optimized with the BC co-training loss. The BC co-training loss is a popular approach for cross-embodiment policy learning, where a standard Behavior Cloning (BC) loss is computed on the aggregated human and robot dataset. \[
\mathcal{L}_{\text{BC-cotrain}}(\phi, \theta) = \mathbb{E}_{(o,a) \sim D_H \cup D_R} [\mathcal{L_{\text{BC}}}(\pi_\theta(f_\phi(o)), a)]
\]
In practice, per training step, for each embodiment $e \in \{\text{robot}, \text{human}\}$, we compute the conditional flow matching (CFM) loss on a mini-batch of human and robot samples:
\[
    \mathcal{L}_{\text{BC-cotrain}} = \mathcal{L}_{\text{CFM}}^{\text{robot}} + \mathcal{L}_{\text{CFM}}^{\text{human}}
\]
The policy architecture is visualized in Fig.~\ref{fig:arch}. Further details on the policy architecture and algorithm are included in Appendix.

\subsection{Evaluation Setup}
\label{ssec:eval_setup}

\begin{figure}[h]
\centering
\includegraphics[width=\linewidth]{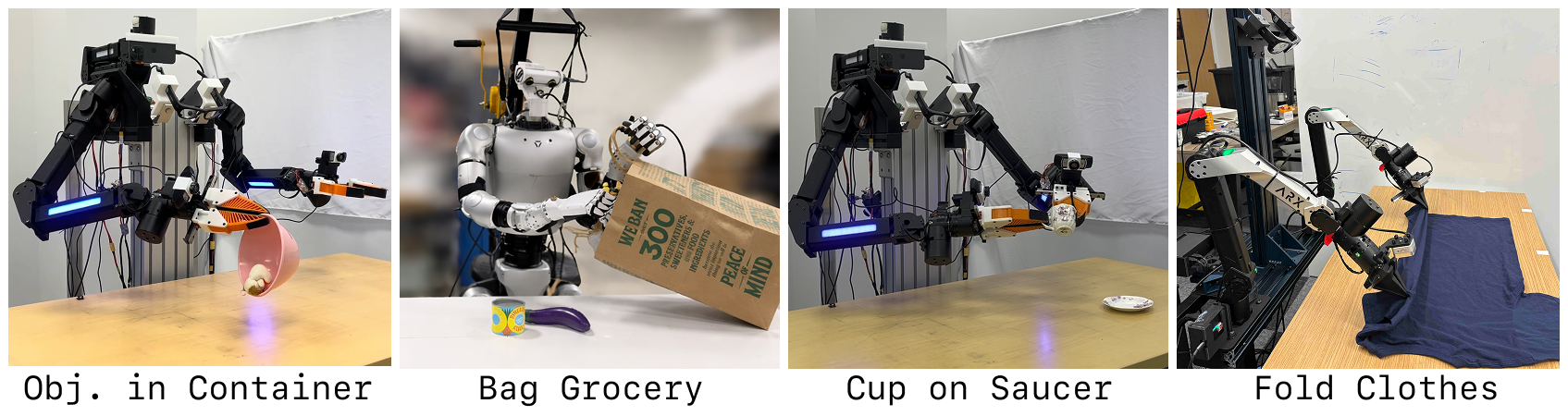}
\caption{\textbf{Evaluation Tasks.} We conduct evaluation with 4 representative Flagship tasks.}
\vspace{-15pt}
\label{fig:tasks}
\end{figure}

\noindent\textbf{Rollout evaluation protocol.}
Evaluation is performed on four representative Flagship tasks shown in Fig.~\ref{fig:tasks}. We evaluate both in-domain (ID) settings, where task layouts match robot training data, and out-of-domain (OOD) settings with unseen objects and environments. For each method, we perform 20 ID and 20 OOD rollouts per task, with randomized initial conditions. Performance is measured using task-specific sub-task metrics, including grasps, placements, intermediate manipulations, and full task completion. For ease of comparison, we report a \textbf{normalized score} aggregated across rollouts. Full task-specific protocols, rollout counts, scoring definitions, and detailed results are provided in the appendix.

\noindent\textbf{Robot data collection.} For each task, approximately 150-300 demonstrations are collected with randomized object placement, orientation and combinations across the workspace. For each task, 4-8 objects are selected from those recorded within \egoversea to be used for demonstrations. Task specific details are provided in the appendix.

\label{ssec:dataset:robots}

\begin{figure}[t]
\centering
\includegraphics[width=0.9\linewidth]{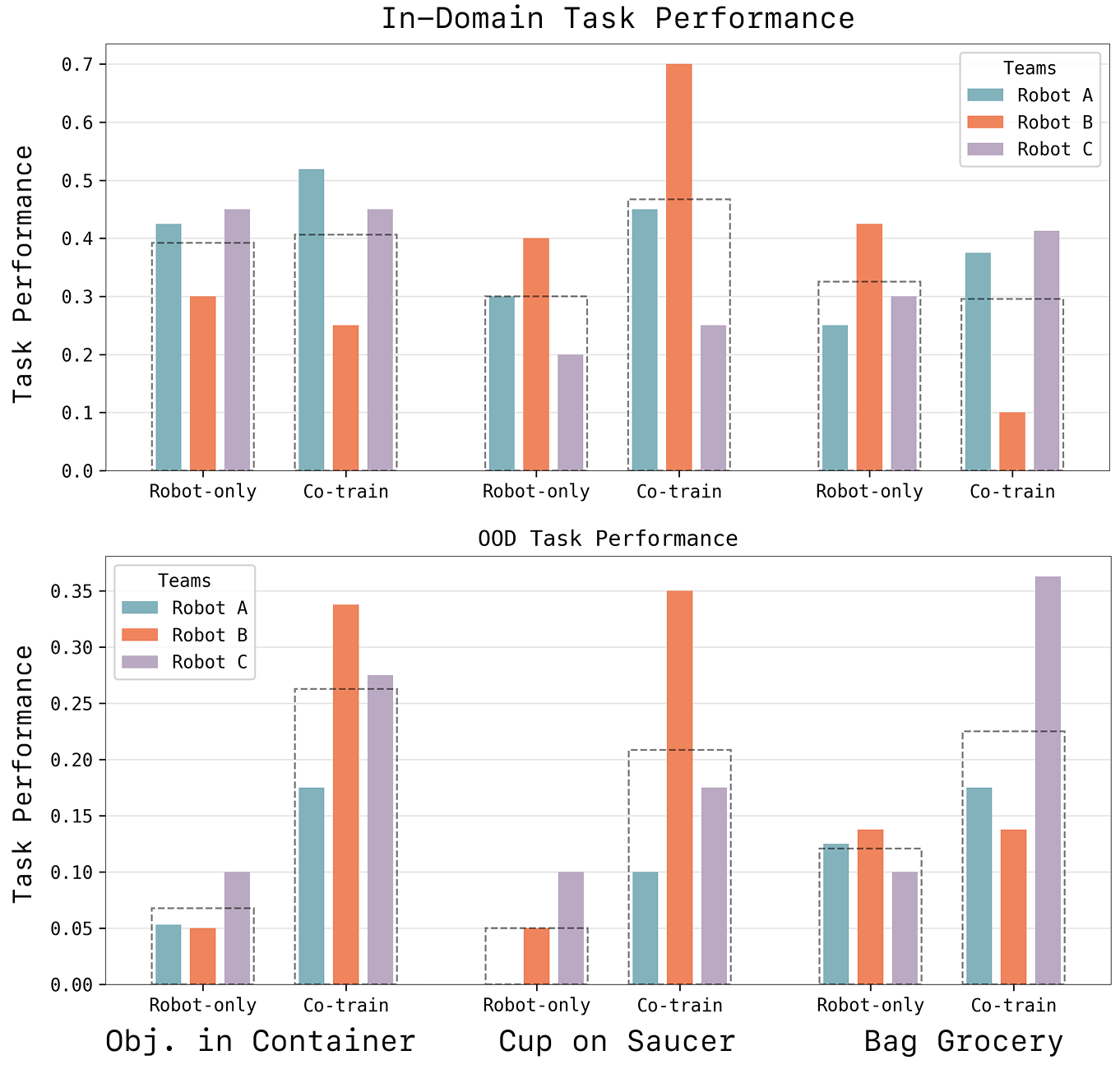}
\caption{\textbf{Co-training improves transfer.} Joint training with human egocentric data consistently improves in-domain performance and out-of-domain generalization across robots.}
\label{fig:flagship}
\vspace{-10pt}
\end{figure}

\begin{figure*}[t]
\centering
\includegraphics[width=0.9\linewidth]{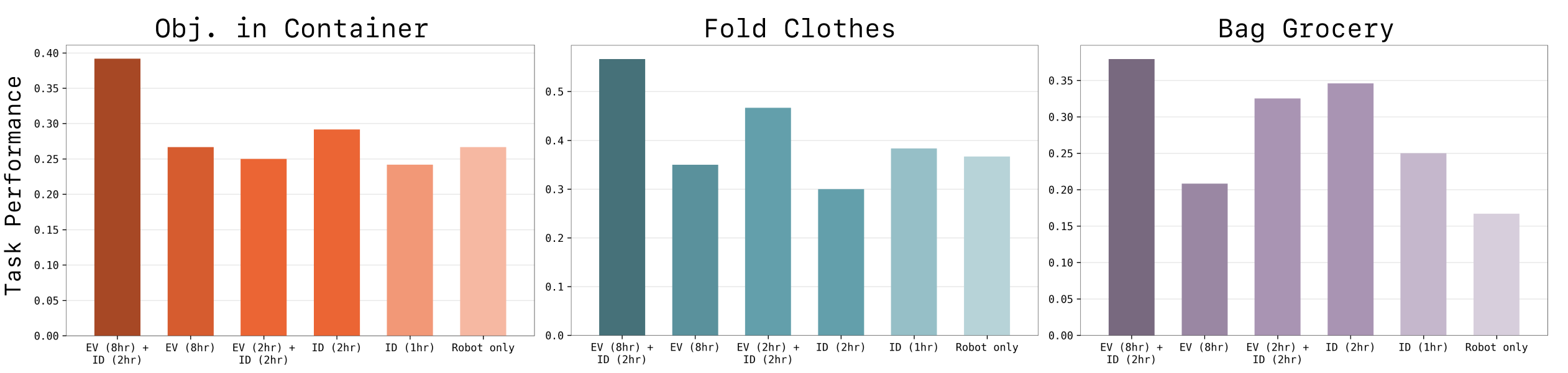}
\caption{\textbf{Domain-aligned data enables scaling.} We ablate the effect of \egoversea (EV) and aligned human data (ID). A small amount of aligned human data anchors learning and allows performance to improve as diverse human data scale.}
\label{fig:scaling}
\vspace{-10pt}
\end{figure*}

\subsection{Does Human Data Scale Robot Performance?}
\label{sec:experiment:realworld}
We investigate whether human egocentric data can reliably improve robot learning and if performance scales as human data volume increases. Rather than optimizing for a single system, we test reproducibility across multiple robots, labs, and three flagship tasks: \emph{object-in-container} (single-arm), \emph{cup-on-saucer} (fine-grained bimanual), and \emph{bag-grocery} (long-horizon bimanual). We co-train all models with a subset of \egoversea data which includes diverse task-specific data and in-domain human data. In-domain human data is collected under the same task definitions as robot teleoperation, but differs in embodiment, sensing, and motion execution. Complete quantitative results are summarized in Fig.~\ref{fig:flagship}.

\textbf{Co-training with \egoversea improves robot performance and generalization.} The presence of \egoversea data consistently improves performance in both in-domain and out-of-domain (OOD) settings by up to $30\%$. It is important to note that the demonstration speed and reset range is different across each manipulation setting. Hence, it is more informative to compare relative improvements from adding human data as opposed to absolute scores. While results are robust across most embodiments and tasks, we observed a performance decrease in the \emph{bag-grocery} task for \robotStanford, whereas \robotGT and \robotUCSD saw improvements. We hypothesize that this divergence stems from \robotStanford's specific embodiment limitations, which forced robot demonstrations to deviate from the human strategies in \egoversea. We include additional analysis of this observation in the Appendix.

\textbf{Domain-aligned data enables transfer from diverse data. } We additionally study whether robot performance scales with diverse human data for each task from \egoversea. We scale the in-domain human data and diverse human data, while the robot data is kept fixed. For scaling to be effective, the policy must extract transferable structure from diverse human behavior. As shown in Fig.~\ref{fig:scaling}, we find that scaling benefits depend critically on the availability of aligned human–robot data. While neither 8h of diverse \egoversea data nor domain-aligned human data alone is sufficient to drive significant performance gains in ID or OOD settings, we observe positive scaling when domain-aligned data ``anchors'' the learning process. This effect enables the policy to effectively transfer knowledge from diverse sources; for instance, the inclusion of just 2h of domain-aligned data facilitates transfer from 2h of diverse \egoversea data, a trend that scales further as the diverse data volume increases to 8h. 

\subsection{How Does Human Data Diversity Affect Generalization?}
\label{sec:experiment:diversity}
\begin{figure}[t]
\centering
\includegraphics[width=1.0\linewidth]{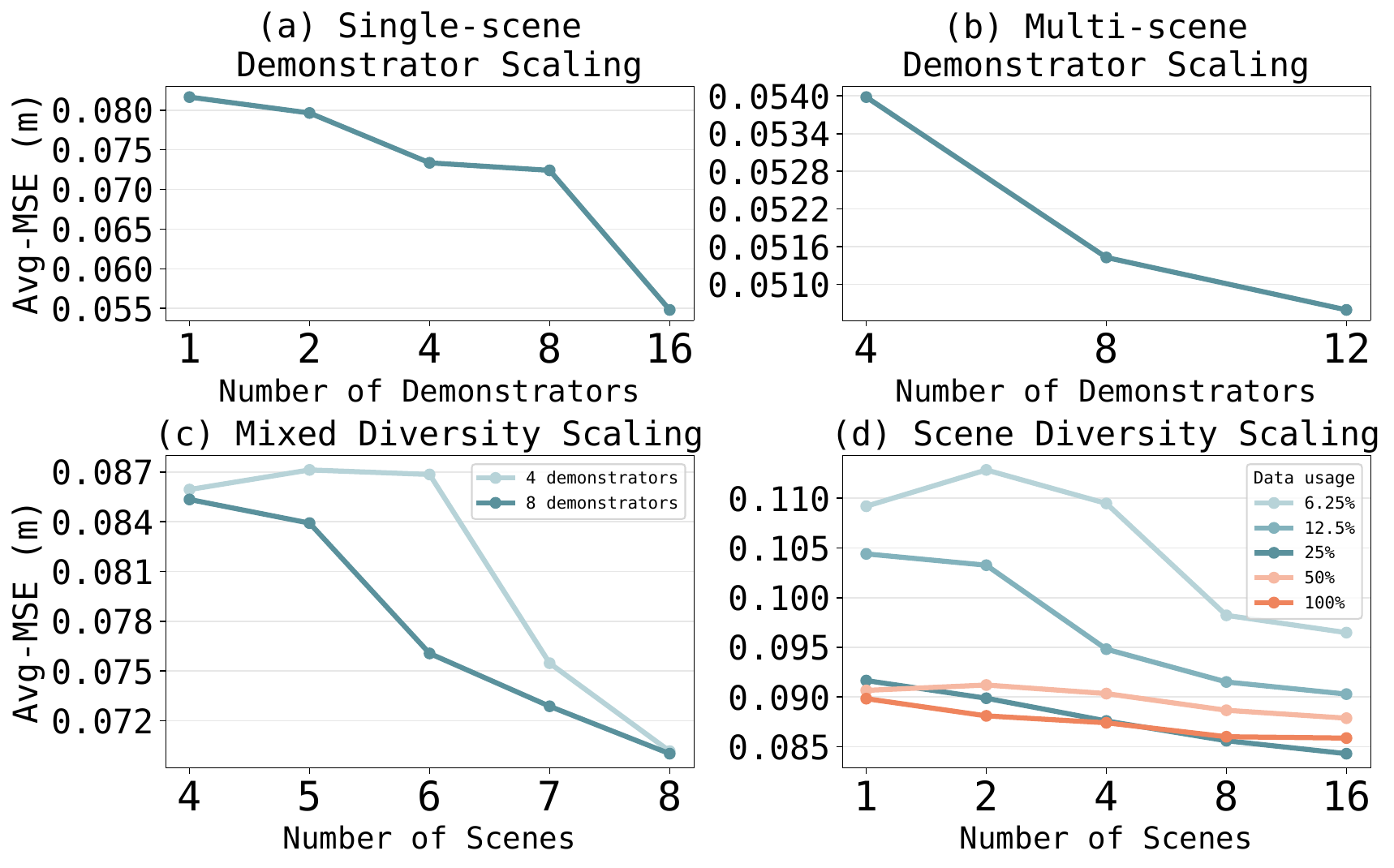}
\caption{\textbf{Controlled Diversity Results.} 
(a) Scaling demonstrators in a single scene improves generalization to unseen ones. 
(b) Demonstrator scaling remains beneficial across eight fixed scenes.
(c) Jointly scaling scene and demonstrator diversity yields complementary improvements.
(d) Increasing scene diversity improves generalization to unseen scenes across data budgets, with the strongest gains under limited data. }
\label{fig:diversity_exp}
\vspace{-10pt}
\end{figure}

\begin{figure}[t]
    \centering
    \includegraphics[width=0.8\linewidth]{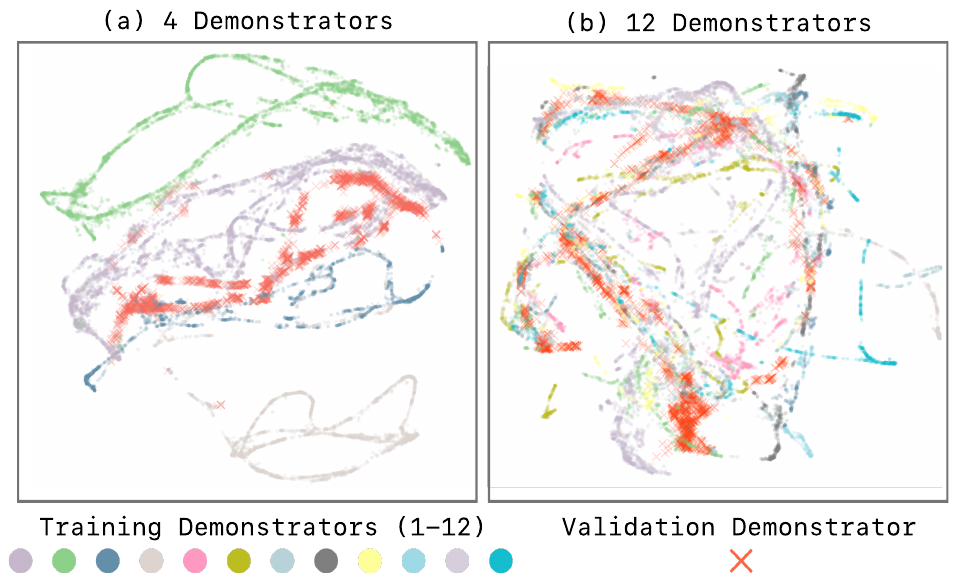}
    \caption{\textbf{Demonstrator Diversity Visualization.} We visualize UMAP embeddings of encoded features for 4 and 12 demonstrators in the multi-scene demonstrator scaling setting, showing greater overlap between training and validation demonstrators with increased demonstrator diversity. }
    \label{fig:diversity_tsne}
    \vspace{-15pt}
\end{figure}


The experiments above rely on naturally aggregated data from multiple labs, which introduces uncontrolled and potentially imbalanced diversity. Prior work has shown that different sources of diversity in robot datasets, such as environments and viewpoints, can have uneven effects on generalization~\cite{khazatsky2024droid, hu2024data, saxena2025whatmatters, zha2025guidingdatacollectionfactored}. Motivated by these findings, we conduct controlled studies to isolate the effects of scene and demonstrator diversity in a human data collection setting described in Sec.~\ref{ssec:dataset:protocol}.
We report primarily the offline Avg-MSE metric in the human-based evaluation setting. While this metric does not directly measure downstream robot performance, it provides a stable signal for comparing generalization across diversity and data scaling regimes~\cite{generalist2025gen0}. Fig.~\ref{fig:diversity_exp} reports results for the \textit{fold-clothes} task, with additional results in Appendix.

\noindent\textbf{Demonstrator Diversity Scaling.}
We vary the number of human demonstrators while controlling for task and scene. We study two settings: (1) \textit{Single scene}, where models are trained with $\{1,2,4,8,16\}$ demonstrators under a fixed 2-hour budget and evaluated on a held-out demonstrator in the same scene; and (2) \textit{Multi scene}, where models are trained with $\{4,8,12\}$ demonstrators under a fixed 8-hour budget across 8 scenes and evaluated on held-out demonstrators in the same environments. As shown in Fig.~\ref{fig:diversity_exp}(a,b), increasing demonstrator diversity consistently improves generalization to unseen demonstrators. To illustrate this effect, Fig.~\ref{fig:diversity_tsne} visualizes UMAP embeddings of encoded features for $\{4, 12\}$ demonstrators, showing increased overlap between training and validation demonstrators with greater demonstrator diversity.



\noindent\textbf{Scene Diversity Scaling.}
We vary the number of training scenes in $\{1,2,4,8,16\}$ while holding the demonstrator pool fixed, and evaluate on unseen scenes. As shown in Fig.~\ref{fig:diversity_exp}(d), generalization performance improves as scene diversity increases across data budgets. Once data quantity reaches a moderate level, increasing data density yields diminishing returns, whereas expanding scene coverage provides measurable gains. This suggests that beyond a certain scale, environmental diversity plays a more critical role in generalization than additional data collected within individual scenes.


\noindent\textbf{Multi-Scene Interaction Effects.}
Finally, we jointly scale scene and demonstrator diversity, reflecting realistic \egoverse data collection. We vary the number of training scenes in $\{4,5,6,7,8\}$ and demonstrators in $\{4,8\}$ under a fixed 4-hour budget, and evaluate on unseen scenes and demonstrators. As shown in Fig.~\ref{fig:diversity_exp}(c), increasing scene diversity improves generalization under both demonstrator budgets, while the marginal benefit of additional demonstrators decreases as scene coverage grows.

\section{Limitations} 
\label{sec:limitations}
Our study mainly focused on human-and-robot co-training. Future work should conduct broader algorithmic exploration (e.g., pre-train and fine-tune). 
Moreover, the scene and demonstrator diversity experiments rely exclusively on offline metrics. While these metrics capture generalization across human demonstrators and environments, additional robot rollouts are necessary to determine whether the observed diversity effects translate to improved generalization in robotic manipulation.

\section{Conclusion} 
\label{sec:conclusion}
In this work, we introduced \egoverse, a collaborative framework for scalable human data-driven robot learning. Our paper presents a large-scale human dataset collected across academic and industry partners, with unprecedented diversity in task semantics, scenarios, and demonstrators. Unlike prior static datasets, \egoverse is designed as a continuously growing resource that supports reproducible study. Beyond the dataset, our consortium-scale evaluation shows that human data improves robot performance, that aligned human–robot data are necessary to anchor effective scaling, and that scene diversity strongly affects generalization under limited data budgets.
\section{Contributions}

\noindent\textbf{Project Leads:}
Ryan Punamiya, Simar Kareer.

\noindent\textbf{Lab Leads (coordinated major components of the project including
data collection pipelines, infrastructure, experiments, and cross-team coordination):}
Zeyi Liu, Josh Citron,
Ri-Zhao Qiu, Xiongyi Cai,
Alexey Gavryushin, Jiaqi Chen, Davide Liconti.

\noindent\textbf{Core Contributors (made substantive contributions to the scientific outcome
including data collection, experiments, system development, analysis, and writing):}
Lawrence Y.\ Zhu, Patcharapong Aphiwetsa, Baoyu Li, Aniketh Chuleva,
Pranav Kuppili, Yangcen Liu, Dhruv Patel, Aidan Gao,
Hye-Young Chung, Ryan Co,
Renee Zbizika, Jeff Liu, Xiaomeng Xu, Haoyu Xiong,
Geng Chen,
Sebastiano Oliani, Wenkai Xuan, Chenyu Yang, Xi Wang.

\noindent\textbf{Industry Partners (contributed engineering support, infrastructure,
and dataset collaboration):}
James Fort, Richard Newcombe,
Josh Gao, Jason Chong,
Garrett Matsuda, Aseem Doriwala.

\noindent\textbf{Academic PIs (provided project supervision and research direction):}
Marc Pollefeys, Robert Katzschmann,
Xiaolong Wang,
Shuran Song,
Judy Hoffman, Danfei Xu.

\vspace{0.75em}

We thank additional collaborators who supported data collection
efforts in participating labs.

\noindent\textbf{Georgia Tech Data Collection:} Zhenyang Chen, Woo Chul Shin, Shuo Cheng, Liqian Ma, Xinchen Yin, Rohan Bansal, David He, Vaibhav Saxena, Mengying Lin, Nadun Ranawaka

\noindent\textbf{ETH Zürich Data Collection:}
Aristotelis Sympetheros, Esteban Padilla Cerdio,
Filippos Katsimalis, Robert Jomar Malate.

\noindent\textbf{Industry Support:}
We thank collaborators at Meta Reality Labs Research, Mecka AI, and Scale AI for data contribution, engineering support,
infrastructure assistance, and collaboration on dataset development.



\bibliographystyle{plainnat}
\bibliography{references}

@article{via,
  title = {Vision in Action: Learning Active Perception from Human Demonstrations},
  author = {Haoyu Xiong and Xiaomeng Xu and Jimmy Wu and Yifan Hou and Jeannette Bohg and Shuran Song},
  journal = {arXiv preprint arXiv:2506.15666},
  year = {2025}
}

@article{avid,
  title={Avid: Learning multi-stage tasks via pixel-level translation of human videos},
  author={Smith, Laura and Dhawan, Nikita and Zhang, Marvin and Abbeel, Pieter and Levine, Sergey},
  journal={arXiv preprint arXiv:1912.04443},
  year={2019}
}

@article{egozero,
  title={Egozero: Robot learning from smart glasses},
  author={Liu, Vincent and Adeniji, Ademi and Zhan, Haotian and Haldar, Siddhant and Bhirangi, Raunaq and Abbeel, Pieter and Pinto, Lerrel},
  journal={arXiv preprint arXiv:2505.20290},
  year={2025}
}

@article{fang2023rh20t,
  title={Rh20t: A comprehensive robotic dataset for learning diverse skills in one-shot},
  author={Fang, Hao-Shu and Fang, Hongjie and Tang, Zhenyu and Liu, Jirong and Wang, Chenxi and Wang, Junbo and Zhu, Haoyi and Lu, Cewu},
  journal={arXiv preprint arXiv:2307.00595},
  year={2023}
}

@InProceedings{hoi4d,
    author    = {Liu, Yunze and Liu, Yun and Jiang, Che and Lyu, Kangbo and Wan, Weikang and Shen, Hao and Liang, Boqiang and Fu, Zhoujie and Wang, He and Yi, Li},
    title     = {HOI4D: A 4D Egocentric Dataset for Category-Level Human-Object Interaction},
    booktitle = {Proceedings of the IEEE/CVF Conference on Computer Vision and Pattern Recognition (CVPR)},
    month     = {June},
    year      = {2022},
    pages     = {21013-21022}
}

@article{UniPi,
  title={Learning universal policies via text-guided video generation},
  author={Du, Yilun and Yang, Sherry and Dai, Bo and Dai, Hanjun and Nachum, Ofir and Tenenbaum, Josh and Schuurmans, Dale and Abbeel, Pieter},
  journal={Advances in neural information processing systems},
  volume={36},
  pages={9156--9172},
  year={2023}
}

@article{egodex,
  title={EgoDex: Learning Dexterous Manipulation from Large-Scale Egocentric Video},
  author={Hoque, Ryan and Huang, Peide and Yoon, David J and Sivapurapu, Mouli and Zhang, Jian},
  journal={arXiv preprint arXiv:2505.11709},
  year={2025}
}

@article{hot3d,
  title={Introducing HOT3D: An egocentric dataset for 3D hand and object tracking},
  author={Banerjee, Prithviraj and Shkodrani, Sindi and Moulon, Pierre and Hampali, Shreyas and Zhang, Fan and Fountain, Jade and Miller, Edward and Basol, Selen and Newcombe, Richard and Wang, Robert and others},
  journal={arXiv preprint arXiv:2406.09598},
  year={2024}
}

@INPROCEEDINGS{lbw,
  author={Xiong, Haoyu and Li, Quanzhou and Chen, Yun-Chun and Bharadhwaj, Homanga and Sinha, Samarth and Garg, Animesh},
  booktitle={2021 IEEE/RSJ International Conference on Intelligent Robots and Systems (IROS)}, 
  title={Learning by Watching: Physical Imitation of Manipulation Skills from Human Videos}, 
  year={2021},
  volume={},
  number={},
  pages={7827-7834},
  doi={10.1109/IROS51168.2021.9636080}}

@inproceedings{Epic-Kitchens,
  title={Scaling egocentric vision: The epic-kitchens dataset},
  author={Damen, Dima and Doughty, Hazel and Farinella, Giovanni Maria and Fidler, Sanja and Furnari, Antonino and Kazakos, Evangelos and Moltisanti, Davide and Munro, Jonathan and Perrett, Toby and Price, Will and others},
  booktitle={Proceedings of the European conference on computer vision (ECCV)},
  pages={720--736},
  year={2018}
}

@inproceedings{EgoExo4D,
  title={Ego-exo4d: Understanding skilled human activity from first-and third-person perspectives},
  author={Grauman, Kristen and Westbury, Andrew and Torresani, Lorenzo and Kitani, Kris and Malik, Jitendra and Afouras, Triantafyllos and Ashutosh, Kumar and Baiyya, Vijay and Bansal, Siddhant and Boote, Bikram and others},
  booktitle={Proceedings of the IEEE/CVF Conference on Computer Vision and Pattern Recognition},
  pages={19383--19400},
  year={2024}
}

@InProceedings{sth-sth,
author = {Goyal, Raghav and Ebrahimi Kahou, Samira and Michalski, Vincent and Materzynska, Joanna and Westphal, Susanne and Kim, Heuna and Haenel, Valentin and Fruend, Ingo and Yianilos, Peter and Mueller-Freitag, Moritz and Hoppe, Florian and Thurau, Christian and Bax, Ingo and Memisevic, Roland},
title = {The "Something Something" Video Database for Learning and Evaluating Visual Common Sense},
booktitle = {Proceedings of the IEEE International Conference on Computer Vision (ICCV)},
month = {Oct},
year = {2017}
}

@misc{kareer2024egomimicscalingimitationlearning,
      title={EgoMimic: Scaling Imitation Learning via Egocentric Video}, 
      author={Simar Kareer and Dhruv Patel and Ryan Punamiya and Pranay Mathur and Shuo Cheng and Chen Wang and Judy Hoffman and Danfei Xu},
      year={2024},
      eprint={2410.24221},
      archivePrefix={arXiv},
      primaryClass={cs.RO},
      url={https://arxiv.org/abs/2410.24221}, 
}

@misc{ye2024latentactionpretrainingvideos,
      title={Latent Action Pretraining from Videos}, 
      author={Seonghyeon Ye and Joel Jang and Byeongguk Jeon and Sejune Joo and Jianwei Yang and Baolin Peng and Ajay Mandlekar and Reuben Tan and Yu-Wei Chao and Bill Yuchen Lin and Lars Liden and Kimin Lee and Jianfeng Gao and Luke Zettlemoyer and Dieter Fox and Minjoon Seo},
      year={2024},
      eprint={2410.11758},
      archivePrefix={arXiv},
      primaryClass={cs.RO},
      url={https://arxiv.org/abs/2410.11758}, 
}

@misc{wang2024hpt,
      title={Scaling Proprioceptive-Visual Learning with Heterogeneous Pre-trained Transformers}, 
      author={Lirui Wang and Xinlei Chen and Jialiang Zhao and Kaiming He},
      year={2024},
      eprint={2409.20537},
      archivePrefix={arXiv},
      primaryClass={cs.RO},
      url={https://arxiv.org/abs/2409.20537}, 
}

@misc{nvidia2025gr00tn1openfoundation,
      title={GR00T N1: An Open Foundation Model for Generalist Humanoid Robots}, 
      author={NVIDIA and : and Johan Bjorck and Fernando Castañeda and Nikita Cherniadev and Xingye Da and Runyu Ding and Linxi "Jim" Fan and Yu Fang and Dieter Fox and Fengyuan Hu and Spencer Huang and Joel Jang and Zhenyu Jiang and Jan Kautz and Kaushil Kundalia and Lawrence Lao and Zhiqi Li and Zongyu Lin and Kevin Lin and Guilin Liu and Edith Llontop and Loic Magne and Ajay Mandlekar and Avnish Narayan and Soroush Nasiriany and Scott Reed and You Liang Tan and Guanzhi Wang and Zu Wang and Jing Wang and Qi Wang and Jiannan Xiang and Yuqi Xie and Yinzhen Xu and Zhenjia Xu and Seonghyeon Ye and Zhiding Yu and Ao Zhang and Hao Zhang and Yizhou Zhao and Ruijie Zheng and Yuke Zhu},
      year={2025},
      eprint={2503.14734},
      archivePrefix={arXiv},
      primaryClass={cs.RO},
      url={https://arxiv.org/abs/2503.14734}, 
}

@misc{bahl2022humantorobotimitationwild,
      title={Human-to-Robot Imitation in the Wild}, 
      author={Shikhar Bahl and Abhinav Gupta and Deepak Pathak},
      year={2022},
      eprint={2207.09450},
      archivePrefix={arXiv},
      primaryClass={cs.RO},
      url={https://arxiv.org/abs/2207.09450}, 
}

@article{black2410pi0,
  title={$\pi$0: A vision-language-action flow model for general robot control, 2024},
  year = {2024},
  author={Black, Kevin and Brown, Noah and Driess, Danny and Esmail, Adnan and Equi, Michael and Finn, Chelsea and Fusai, Niccolo and Groom, Lachy and Hausman, Karol and Ichter, Brian and others},
  journal={URL https://arxiv. org/abs/2410.24164}
}

@article{hu2024data,
  title={Data scaling laws in imitation learning for robotic manipulation},
  author={Hu, Yingdong and Lin, Fanqi and Sheng, Pingyue and Wen, Chuan and You, Jiacheng and Gao, Yang},
  journal={arXiv preprint arXiv:2410.18647},
  year={2024}
}

@inproceedings{grauman2022ego4d,
  title={Ego4d: Around the world in 3,000 hours of egocentric video},
  author={Grauman, Kristen and Westbury, Andrew and Byrne, Eugene and Chavis, Zachary and Furnari, Antonino and Girdhar, Rohit and Hamburger, Jackson and Jiang, Hao and Liu, Miao and Liu, Xingyu and others},
  booktitle={Proceedings of the IEEE/CVF conference on computer vision and pattern recognition},
  pages={18995--19012},
  year={2022}
}

@inproceedings{fan2023arctic,
  title={ARCTIC: A dataset for dexterous bimanual hand-object manipulation},
  author={Fan, Zicong and Taheri, Omid and Tzionas, Dimitrios and Kocabas, Muhammed and Kaufmann, Manuel and Black, Michael J and Hilliges, Otmar},
  booktitle={Proceedings of the IEEE/CVF conference on computer vision and pattern recognition},
  pages={12943--12954},
  year={2023}
}

@misc{pi0.5,
      title={$\pi_{0.5}$: a Vision-Language-Action Model with Open-World Generalization}, 
      author={Physical Intelligence and Kevin Black and Noah Brown and James Darpinian and Karan Dhabalia and Danny Driess and Adnan Esmail and Michael Equi and Chelsea Finn and Niccolo Fusai and Manuel Y. Galliker and Dibya Ghosh and Lachy Groom and Karol Hausman and Brian Ichter and Szymon Jakubczak and Tim Jones and Liyiming Ke and Devin LeBlanc and Sergey Levine and Adrian Li-Bell and Mohith Mothukuri and Suraj Nair and Karl Pertsch and Allen Z. Ren and Lucy Xiaoyang Shi and Laura Smith and Jost Tobias Springenberg and Kyle Stachowicz and James Tanner and Quan Vuong and Homer Walke and Anna Walling and Haohuan Wang and Lili Yu and Ury Zhilinsky},
      year={2025},
      eprint={2504.16054},
      archivePrefix={arXiv},
      primaryClass={cs.LG},
      url={https://arxiv.org/abs/2504.16054}, 
}

@article{chi2024diffusionpolicy,
	author = {Cheng Chi and Zhenjia Xu and Siyuan Feng and Eric Cousineau and Yilun Du and Benjamin Burchfiel and Russ Tedrake and Shuran Song},
	title ={Diffusion Policy: Visuomotor Policy Learning via Action Diffusion},
	journal = {The International Journal of Robotics Research},
	year = {2024},
}

@inproceedings{o2024open,
  title={Open x-embodiment: Robotic learning datasets and rt-x models: Open x-embodiment collaboration 0},
  author={O’Neill, Abby and Rehman, Abdul and Maddukuri, Abhiram and Gupta, Abhishek and Padalkar, Abhishek and Lee, Abraham and Pooley, Acorn and Gupta, Agrim and Mandlekar, Ajay and Jain, Ajinkya and others},
  booktitle={2024 IEEE International Conference on Robotics and Automation (ICRA)},
  pages={6892--6903},
  year={2024},
  organization={IEEE}
}

@article{khazatsky2024droid,
  title={Droid: A large-scale in-the-wild robot manipulation dataset},
  author={Khazatsky, Alexander and Pertsch, Karl and Nair, Suraj and Balakrishna, Ashwin and Dasari, Sudeep and Karamcheti, Siddharth and Nasiriany, Soroush and Srirama, Mohan Kumar and Chen, Lawrence Yunliang and Ellis, Kirsty and others},
  journal={arXiv preprint arXiv:2403.12945},
  year={2024}
}

@inproceedings{bharadhwaj2024track2act,
  author    = {Homanga Bharadhwaj and Roozbeh Mottaghi and Abhinav Gupta and Shubham Tulsiani},
  title     = {Track2Act: Predicting Point Tracks from Internet Videos enables Generalizable Robot Manipulation},
  booktitle = {European Conference on Computer Vision (ECCV)},
  year      = {2024}
}

@misc{wen2023anypoint,
      title={Any-point Trajectory Modeling for Policy Learning},
      author={Chuan Wen and Xingyu Lin and John So and Kai Chen and Qi Dou and Yang Gao and Pieter Abbeel},
      year={2023},
      eprint={2401.00025},
      archivePrefix={arXiv},
      primaryClass={cs.RO}
}

@misc{ren2025motiontracksunifiedrepresentation,
      title={Motion Tracks: A Unified Representation for Human-Robot Transfer in Few-Shot Imitation Learning}, 
      author={Juntao Ren and Priya Sundaresan and Dorsa Sadigh and Sanjiban Choudhury and Jeannette Bohg},
      year={2025},
      eprint={2501.06994},
      archivePrefix={arXiv},
      primaryClass={cs.RO},
      url={https://arxiv.org/abs/2501.06994}, 
}

@article{qiu2025-humanpolicy,
    title={Humanoid Policy \~{} Human Policy},
    author={Ri-Zhao Qiu and Shiqi Yang and Xuxin Cheng and Chaitanya Chawla and Jialong Li and Tairan He and Ge Yan and David J. Yoon and Ryan Hoque and Lars Paulsen and Ge Yang and Jian Zhang and Sha Yi and Guanya Shi and Xiaolong Wang},
    journal={arXiv preprint arXiv:2503.13441},
    year={2025}
  }

@misc{lepert2025phantomtrainingrobotsrobots,
      title={Phantom: Training Robots Without Robots Using Only Human Videos}, 
      author={Marion Lepert and Jiaying Fang and Jeannette Bohg},
      year={2025},
      eprint={2503.00779},
      archivePrefix={arXiv},
      primaryClass={cs.RO},
      url={https://arxiv.org/abs/2503.00779}, 
}

@inproceedings{etukuru2025robot,
  title={Robot utility models: General policies for zero-shot deployment in new environments},
  author={Etukuru, Haritheja and Naka, Norihito and Hu, Zijin and Lee, Seungjae and Mehu, Julian and Edsinger, Aaron and Paxton, Chris and Chintala, Soumith and Pinto, Lerrel and Shafiullah, Nur Muhammad Mahi},
  booktitle={2025 IEEE International Conference on Robotics and Automation (ICRA)},
  pages={8275--8283},
  year={2025},
  organization={IEEE}
}

@article{chi2024universal,
  title={Universal manipulation interface: In-the-wild robot teaching without in-the-wild robots},
  author={Chi, Cheng and Xu, Zhenjia and Pan, Chuer and Cousineau, Eric and Burchfiel, Benjamin and Feng, Siyuan and Tedrake, Russ and Song, Shuran},
  journal={arXiv preprint arXiv:2402.10329},
  year={2024}
}

@article{yin2025osmo,
  title={OSMO: Open-Source Tactile Glove for Human-to-Robot Skill Transfer},
  author={Yin, Jessica and Qi, Haozhi and Wi, Youngsun and Kundu, Sayantan and Lambeta, Mike and Yang, William and Wang, Changhao and Wu, Tingfan and Malik, Jitendra and Hellebrekers, Tess},
  journal={arXiv preprint arXiv:2512.08920},
  year={2025}
}

@misc{octomodelteam2024octoopensourcegeneralistrobot,
      title={Octo: An Open-Source Generalist Robot Policy}, 
      author={Octo Model Team and Dibya Ghosh and Homer Walke and Karl Pertsch and Kevin Black and Oier Mees and Sudeep Dasari and Joey Hejna and Tobias Kreiman and Charles Xu and Jianlan Luo and You Liang Tan and Lawrence Yunliang Chen and Pannag Sanketi and Quan Vuong and Ted Xiao and Dorsa Sadigh and Chelsea Finn and Sergey Levine},
      year={2024},
      eprint={2405.12213},
      archivePrefix={arXiv},
      primaryClass={cs.RO},
      url={https://arxiv.org/abs/2405.12213}, 
}

@misc{engel2023projectarianewtool,
      title={Project Aria: A New Tool for Egocentric Multi-Modal AI Research}, 
      author={Jakob Engel and Kiran Somasundaram and Michael Goesele and Albert Sun and Alexander Gamino and Andrew Turner and Arjang Talattof and Arnie Yuan and Bilal Souti and Brighid Meredith and Cheng Peng and Chris Sweeney and Cole Wilson and Dan Barnes and Daniel DeTone and David Caruso and Derek Valleroy and Dinesh Ginjupalli and Duncan Frost and Edward Miller and Elias Mueggler and Evgeniy Oleinik and Fan Zhang and Guruprasad Somasundaram and Gustavo Solaira and Harry Lanaras and Henry Howard-Jenkins and Huixuan Tang and Hyo Jin Kim and Jaime Rivera and Ji Luo and Jing Dong and Julian Straub and Kevin Bailey and Kevin Eckenhoff and Lingni Ma and Luis Pesqueira and Mark Schwesinger and Maurizio Monge and Nan Yang and Nick Charron and Nikhil Raina and Omkar Parkhi and Peter Borschowa and Pierre Moulon and Prince Gupta and Raul Mur-Artal and Robbie Pennington and Sachin Kulkarni and Sagar Miglani and Santosh Gondi and Saransh Solanki and Sean Diener and Shangyi Cheng and Simon Green and Steve Saarinen and Suvam Patra and Tassos Mourikis and Thomas Whelan and Tripti Singh and Vasileios Balntas and Vijay Baiyya and Wilson Dreewes and Xiaqing Pan and Yang Lou and Yipu Zhao and Yusuf Mansour and Yuyang Zou and Zhaoyang Lv and Zijian Wang and Mingfei Yan and Carl Ren and Renzo De Nardi and Richard Newcombe},
      year={2023},
      eprint={2308.13561},
      archivePrefix={arXiv},
      primaryClass={cs.HC},
      url={https://arxiv.org/abs/2308.13561}, 
}

@inproceedings{punamiyaegobridge,
  title={EgoBridge: Domain Adaptation for Generalizable Imitation from Egocentric Human Data},
  author={Punamiya, Ryan and Patel, Dhruv and Aphiwetsa, Patcharapong and Kuppili, Pranav and Zhu, Lawrence Y and Kareer, Simar and Hoffman, Judy and Xu, Danfei},
  booktitle={The Thirty-ninth Annual Conference on Neural Information Processing Systems}
}

@misc{immimic,
      title={ImMimic: Cross-Domain Imitation from Human Videos via Mapping and Interpolation}, 
      author={Yangcen Liu and Woo Chul Shin and Yunhai Han and Zhenyang Chen and Harish Ravichandar and Danfei Xu},
      year={2025},
      eprint={2509.10952},
      archivePrefix={arXiv},
      primaryClass={cs.RO},
      url={https://arxiv.org/abs/2509.10952}, 
}

@inproceedings{saxena2025whatmatters,
    author = {Vaibhav Saxena and Matthew Bronars and Nadun Ranawaka Arachchige and Kuancheng Wang and Woo Chul Shin and Soroush Nasiriany and Ajay Mandlekar and Danfei Xu},
    title = {What Matters in Learning from Large-Scale Datasets for Robot Manipulation},
    booktitle = {The Thirteenth International Conference on Learning Representations},
    year = {2025}
}

@misc{zha2025guidingdatacollectionfactored,
        title={Guiding Data Collection via Factored Scaling Curves}, 
        author={Lihan Zha and Apurva Badithela and Michael Zhang and Justin Lidard and Jeremy Bao and Emily Zhou and David Snyder and Allen Z. Ren and Dhruv Shah and Anirudha Majumdar},
        year={2025},
        eprint={2505.07728},
        archivePrefix={arXiv},
        primaryClass={cs.RO},
        url={https://arxiv.org/abs/2505.07728}, 
}

@article{zhu2026emma,
  author={Zhu, Lawrence Y. and Kuppili, Pranav and Punamiya, Ryan and Aphiwetsa, Patcharapong and Patel, Dhruv and Kareer, Simar and Ha, Sehoon and Xu, Danfei},
  journal={IEEE Robotics and Automation Letters}, 
  title={EMMA: Scaling Mobile Manipulation via Egocentric Human Data}, 
  year={2026},
  volume={11},
  number={3},
  pages={3087-3094},
  doi={10.1109/LRA.2026.3653320}}

@article{generalist2025gen0,
author = {Generalist AI Team},
title = {GEN-0: Embodied Foundation Models That Scale with Physical Interaction},
journal = {Generalist AI Blog},
year = {2025},
note = {https://generalistai.com/blog/nov-04-2025-GEN-0},
}

@article{bu2025univla,
  title={Univla: Learning to act anywhere with task-centric latent actions},
  author={Bu, Qingwen and Yang, Yanting and Cai, Jisong and Gao, Shenyuan and Ren, Guanghui and Yao, Maoqing and Luo, Ping and Li, Hongyang},
  journal={arXiv preprint arXiv:2505.06111},
  year={2025}
}

@misc{chen2025largevideoplanner,
  title={Large Video Planner}, 
  author={Boyuan Chen and Tianyuan Zhang and Haoran Geng and Kiwhan Song and William T. Freeman and Jitendra Malik and Russ Tedrake and Vincent Sitzmann and Yilun Du},
  year={2025},
  eprint={2512.15840},
  archivePrefix={arXiv},
  primaryClass={cs.LG},
  url={http://arxiv.org/abs/2512.15840}, 
}

@article{cai2025n,
  title   = {In-N-On: Scaling Egocentric Manipulation with in-the-wild and on-task Data},
  author  = {Cai, Xiongyi and Qiu, Ri-Zhao and Chen, Geng and Wei, Lai and Liu, Isabella 
            and Huang, Tianshu and Cheng, Xuxin and Wang, Xiaolong},
  journal = {arXiv preprint arXiv:2511.15704},
  year    = {2025}
}

@misc{kareer2025emergencehumanrobottransfer,
      title={Emergence of Human to Robot Transfer in Vision-Language-Action Models}, 
      author={Simar Kareer and Karl Pertsch and James Darpinian and Judy Hoffman and Danfei Xu and Sergey Levine and Chelsea Finn and Suraj Nair},
      year={2025},
      eprint={2512.22414},
      archivePrefix={arXiv},
      primaryClass={cs.RO},
      url={https://arxiv.org/abs/2512.22414}, 
}

@misc{goswami2025dexwm,
  title={World Models Can Leverage Human Videos for Dexterous Manipulation},
  author={Raktim Gautam Goswami and Amir Bar and David Fan and Tsung-Yen Yang and Gaoyue Zhou and Prashanth Krishnamurthy and Michael Rabbat and Farshad Khorrami and Yann LeCun},
  year={2025},
  eprint={2512.13644},
  archivePrefix={arXiv},
  primaryClass={cs.RO},
  url={https://arxiv.org/abs/2512.13644},
}

@misc{yang2025egovlalearningvisionlanguageactionmodels,
        title={EgoVLA: Learning Vision-Language-Action Models from Egocentric Human Videos}, 
        author={Ruihan Yang and Qinxi Yu and Yecheng Wu and Rui Yan and Borui Li and An-Chieh Cheng and Xueyan Zou and Yunhao Fang and Hongxu Yin and Sifei Liu and Song Han and Yao Lu and Xiaolong Wang},
        year={2025},
        eprint={2507.12440},
        archivePrefix={arXiv},
        primaryClass={cs.RO},
        url={https://arxiv.org/abs/2507.12440}, 
}

@inproceedings{shi2025zeromimic,
      title={ZeroMimic: Distilling Robotic Manipulation Skills from Web Videos}, 
      author={Junyao Shi and Zhuolun Zhao and Tianyou Wang and Ian Pedroza and Amy Luo and Jie Wang and Jason Ma and Dinesh Jayaraman},
      year={2025},
      booktitle={International Conference on Robotics and Automation (ICRA)},
}

@article{bahl2023affordances,
      title={Affordances from Human Videos as a Versatile Representation for Robotics},
      author={Bahl, Shikhar and Mendonca, Russell and Chen, Lili and Jain, Unnat and Pathak, Deepak},
      journal={CVPR},
      year={2023}
}

@misc{he2025scalingcrossembodimentworldmodels,
      title={Scaling Cross-Embodiment World Models for Dexterous Manipulation}, 
      author={Zihao He and Bo Ai and Tongzhou Mu and Yulin Liu and Weikang Wan and Jiawei Fu and Yilun Du and Henrik I. Christensen and Hao Su},
      year={2025},
      eprint={2511.01177},
      archivePrefix={arXiv},
      primaryClass={cs.RO},
      url={https://arxiv.org/abs/2511.01177}, 
}

@article{beingbeyond2025beingh0,
  title={Being-H0: Vision-Language-Action Pretraining from Large-Scale Human Videos},
  author={Luo, Hao and Feng, Yicheng and Zhang, Wanpeng and Zheng, Sipeng and Wang, Ye and Yuan, Haoqi and Liu, Jiazheng and Xu, Chaoyi and Jin, Qin and Lu, Zongqing},
  journal={arXiv preprint arXiv:2507.15597},
  year={2025}
}

@misc{guzey2025aina,
      title={Dexterity from Smart Lenses: Multi-Fingered Robot Manipulation with In-the-Wild Human Demonstrations}, 
      author={Irmak Guzey and Haozhi Qi and Julen Urain and Changhao Wang and Jessica Yin and Krishna Bodduluri and Mike Lambeta and Lerrel Pinto and Akshara Rai and Jitendra Malik and Tingfan Wu and Akash Sharma and Homanga Bharadhwaj},
      year={2025},
      eprint={2511.16661},
      archivePrefix={arXiv},
      primaryClass={cs.RO},
      url={https://arxiv.org/abs/2511.16661}, 
}

@misc{he2015deepresiduallearningimage,
      title={Deep Residual Learning for Image Recognition}, 
      author={Kaiming He and Xiangyu Zhang and Shaoqing Ren and Jian Sun},
      year={2015},
      eprint={1512.03385},
      archivePrefix={arXiv},
      primaryClass={cs.CV},
      url={https://arxiv.org/abs/1512.03385}, 
}

@misc{siméoni2025dinov3,
      title={DINOv3}, 
      author={Oriane Siméoni and Huy V. Vo and Maximilian Seitzer and Federico Baldassarre and Maxime Oquab and Cijo Jose and Vasil Khalidov and Marc Szafraniec and Seungeun Yi and Michaël Ramamonjisoa and Francisco Massa and Daniel Haziza and Luca Wehrstedt and Jianyuan Wang and Timothée Darcet and Théo Moutakanni and Leonel Sentana and Claire Roberts and Andrea Vedaldi and Jamie Tolan and John Brandt and Camille Couprie and Julien Mairal and Hervé Jégou and Patrick Labatut and Piotr Bojanowski},
      year={2025},
      eprint={2508.10104},
      archivePrefix={arXiv},
      primaryClass={cs.CV},
      url={https://arxiv.org/abs/2508.10104}, 
}

@misc{OrbikEbert2021OculusReader,
  author = {Jedrzej Orbik, Frederik Ebert},
  title = {Oculus Reader: Robotic Teleoperation Interface},
  year = {2021},
  url = {https://github.com/rail-berkeley/oculus_reader},
  note = {Accessed: YYYY-MM-DD}
}

@software{Zakka_Mink_Python_inverse_2025,
  author = {Zakka, Kevin},
  title = {{Mink: Python inverse kinematics based on MuJoCo}},
  year = {2025},
  month = dec,
  version = {1.0.0},
  url = {https://github.com/kevinzakka/mink},
  license = {Apache-2.0}
}

@misc{wu2023gello,
    title={GELLO: A General, Low-Cost, and Intuitive Teleoperation Framework for Robot Manipulators},
    author={Philipp Wu and Yide Shentu and Zhongke Yi and Xingyu Lin and Pieter Abbeel},
    year={2023},
}

@misc{li2025amoadaptivemotionoptimization,
      title={AMO: Adaptive Motion Optimization for Hyper-Dexterous Humanoid Whole-Body Control}, 
      author={Jialong Li and Xuxin Cheng and Tianshu Huang and Shiqi Yang and Ri-Zhao Qiu and Xiaolong Wang},
      year={2025},
      eprint={2505.03738},
      archivePrefix={arXiv},
      primaryClass={cs.RO},
      url={https://arxiv.org/abs/2505.03738}, 
}

@article{cheng2024tv,
title={Open-TeleVision: Teleoperation with Immersive Active Visual Feedback},
author={Cheng, Xuxin and Li, Jialong and Yang, Shiqi and Yang, Ge and Wang, Xiaolong},
journal={arXiv preprint arXiv:2407.01512},
year={2024}
}

\section{Appendix}
\subsection{Table of Contents}
\label{appendix:A}

\begin{table}[h]
\centering
\small
\begin{tabular}{cl}
\toprule
\textbf{Appendix} & \textbf{Section} \\
\midrule
A & Table of Contents \\

\hyperref[appendix:B]{B} & Extended Discussion \\

\hyperref[appendix:C]{C} & EgoVerse Data Composition \\

\hyperref[appendix:D]{D} & Human Data Collection Setup Detail \\

\hyperref[appendix:E]{E} & Data Capture Hardware Detail \\

\hyperref[appendix:F]{F} & EgoDB Detail \\


\hyperref[appendix:G]{G} & Data Alignment and Post Processing \\

\hyperref[appendix:H]{H} & Robot Data Collection Setup Detail \\

\hyperref[appendix:I]{I} & Policy Architecture and Learning Detail \\


\hyperref[appendix:J]{J} & Robot Experiment Results \\

\hyperref[appendix:K]{K} & Controlled Diversity Experiments \\

\hyperref[appendix:L]{L} & Latent Space Visualization \\
\bottomrule
\end{tabular}
\caption{Appendix table of contents with section links.}
\label{tab:appendix_toc}
\end{table}
 
\subsection{Extended Discussion}
\label{appendix:B}

This work explores one concrete instantiation of learning from egocentric human data via joint human–robot co-training, but more importantly, it opens a broader design space for future research enabled by \egoverse. A natural next direction is to systematically study different training paradigms beyond co-training, including large-scale pretraining on heterogeneous human data followed by targeted fine-tuning on small amounts of robot or aligned human–robot data. Understanding when pretraining yields transferable structure, and how embodiment-specific design decisions impact transfer, remains an open question that \egoverse is well positioned to support.

Another promising avenue is leveraging the full breadth of unaligned, in-the-wild human data in \egoversei. While our study focuses on more controlled \egoversea, \egoversei contains rich, open-ended manipulation data with dense language annotations. This creates opportunities for language-conditioned or goal-conditioned policies that can learn from broader task distributions without requiring strict alignment at training time. Developing methods that can exploit such weakly aligned or unaligned data, while still grounding execution through limited aligned supervision, is a key step toward scalable human-to-robot transfer.

Finally, the consortium-scale nature of \egoverse enables deeper analysis of embodiment effects that go beyond aggregate performance trends. Future work can examine how specific embodiment factors, such as kinematic structure (e.g., gripper vs. dexterous hands), sensing configuration, or control parameterization, interact with different types of human data. Such analyses can inform principled data selection, curriculum design, and model architectures as datasets continue to grow. Taken together, \egoverse is not only a dataset but a shared experimental substrate for studying how representation learning, embodiment, and data scale jointly shape the next generation of generalizable robot policies.

\subsection{\egoverse Data Composition}
\label{appendix:C}

 The amount of data contributed by each partner for \egoversea and \egoversei is summarized in Table~\ref{tab:egoverse_composition}. While providing a per-task breakdown of hours across the full \egoverse dataset is infeasible, we instead report aggregate statistics over semantic task categories and their corresponding dataset sizes in Table~\ref{tab:category_hours}. In addition, we provide a frequency distribution for the top 20 ``verbs" across each category in Table~\ref{tab:category_verbs}.

\begin{table}[t]
\centering
\begin{tabular}{lccrr}
\toprule
\textbf{Part} & \textbf{Percent} & \textbf{\# Hours} & \textbf{\# Episodes} & \textbf{\# Tasks} \\
\midrule
\egoversea  (all partners)            & 5.5\%  & 75    & 2{,}385  & 6 \\
\egoversei partner A    & 76.1\% & 1{,}035 & 72{,}993 & 1{,}898 \\
\egoversei partner B    & 18.4\% & 250   & 3{,}128  & 45 \\
\bottomrule
\end{tabular}
\caption{Dataset composition across EgoVerse components.}
\label{tab:egoverse_composition}
\end{table}

\begin{table}[t]
\centering
\begin{tabular}{lcc}
\toprule
\textbf{Category} & \textbf{Percent} & \textbf{\# Hours} \\
\midrule
Logistics  & 15.4\% & 209 \\
Cooking    & 13.7\% & 186 \\
Cleaning   & 11.6\% & 158 \\
Laundry    & 10.9\% & 148 \\
Hardware   & 6.8\%  & 92  \\
Crafts     & 4.0\%  & 54  \\
Gardening  & 3.2\%  & 44  \\
\bottomrule
\end{tabular}
\caption{Task category distribution.}
\label{tab:category_hours}
\end{table}
\begin{table*}[t]
\centering
\small
\setlength{\tabcolsep}{3pt}
\renewcommand{\arraystretch}{0.95}
\begin{tabular}{|l r|l r|l r|l r|l r|l r|}
\toprule
\textbf{Logistics} & \textbf{Freq.} &
\textbf{Cooking} & \textbf{Freq.} &
\textbf{Cleaning} & \textbf{Freq.} &
\textbf{Laundry} & \textbf{Freq.} &
\textbf{Hardware} & \textbf{Freq.} &
\textbf{Crafts} & \textbf{Freq.} \\
\midrule
pick & 34{,}524 & pick & 16{,}197 & scrub & 20{,}320 & pick & 6{,}998 & pick & 16{,}387 & pick & 7{,}577 \\
scoop & 12{,}164 & place & 6{,}678 & pick & 12{,}297 & fold & 6{,}190 & place & 5{,}476 & place & 2{,}517 \\
place & 10{,}322 & cut & 5{,}975 & clean & 7{,}713 & iron & 5{,}342 & adjust & 5{,}263 & adjust & 1{,}835 \\
fill & 7{,}991 & scoop & 4{,}722 & wipe & 6{,}863 & adjust & 3{,}628 & remove & 3{,}305 & fold & 1{,}059 \\
adjust & 7{,}802 & adjust & 4{,}496 & dip & 5{,}951 & place & 3{,}325 & unscrew & 3{,}222 & cut & 896 \\
seal & 4{,}249 & fill & 2{,}654 & place & 5{,}480 & smooth & 2{,}429 & hold & 2{,}104 & hold & 589 \\
open & 3{,}392 & hold & 2{,}153 & adjust & 5{,}225 & straighten & 1{,}161 & tighten & 2{,}052 & move & 467 \\
put & 3{,}064 & slice & 1{,}974 & hold & 3{,}358 & flip & 763 & clean & 1{,}564 & apply & 434 \\
hold & 3{,}041 & remove & 1{,}687 & remove & 2{,}862 & smoothen & 751 & test & 1{,}253 & attach & 391 \\
insert & 1{,}996 & press & 1{,}646 & wash & 2{,}301 & hold & 560 & put & 1{,}196 & press & 367 \\
close & 1{,}426 & put & 1{,}215 & rinse & 1{,}901 & clean & 487 & scrape & 1{,}068 & align & 352 \\
fold & 1{,}184 & move & 1{,}088 & scrape & 1{,}877 & grab & 463 & fill & 1{,}047 & put & 347 \\
transfer & 1{,}118 & open & 1{,}066 & brush & 1{,}752 & spread & 439 & solder & 908 & wrap & 316 \\
move & 1{,}117 & seal & 1{,}060 & thread & 1{,}647 & move & 372 & move & 786 & shake & 269 \\
clean & 868 & trim & 964 & put & 850 & button & 349 & install & 647 & tie & 239 \\
pack & 760 & flatten & 803 & pull & 668 & lift & 321 & insert & 620 & smooth & 235 \\
wipe & 749 & chop & 729 & tie & 607 & unfold & 314 & apply & 585 & shape & 234 \\
pour & 686 & transfer & 673 & polish & 576 & insert & 300 & screw & 574 & twist & 216 \\
remove & 611 & fold & 632 & move & 558 & press & 295 & open & 569 & remove & 214 \\
add & 586 & knead & 624 & rotate & 524 & trim & 289 & secure & 568 & insert & 200 \\
\bottomrule
\end{tabular}
\caption{Top 20 verbs per category with frequencies. Each row corresponds to the same rank across six task categories.}
\label{tab:category_verbs}
\end{table*}

\subsection{Human Data Collection Setup Detail}
\label{appendix:D}
\egoversea is designed to capture diversity while maintaining controlled task semantics and data quality.

\emph{Scenario and object diversity.}
Each flagship task is performed across 8--12 unique scenes per lab, with 1--10 dataset units collected per scene to capture within-environment variation. Within each scene, a roughly $40\text{cm} \times 60\text{cm}$ workspace is used, and collectors are explicitly encouraged to randomize object positions across demonstrations. Within each lab, objects are sampled from a fixed set of up to 30 objects per task. As objects are procured independently across participating sites, the dataset exhibits broad variation in real-world object geometry, appearance, and material properties.

\emph{Demonstrator diversity.}
Data are collected from 1--8 demonstrators per lab. Even under identical instructions and tutorial videos, demonstrators exhibit distinct motion habits, timing, coordination strategies, and hand trajectories. The dataset therefore captures substantial variation in human morphology and egocentric viewpoints, arising naturally from differences in collector height, posture, and workspace configuration (e.g., seated versus standing).

Prior work has shown that such human motion variability can significantly affect policy learning behavior and cross-embodiment alignment~\cite{kareer2024egomimicscalingimitationlearning, punamiyaegobridge, immimic}. Rather than eliminating this variation, we treat it as an inherent component of human data that must be managed as datasets scale.
While all participating labs follow the same instruction protocol and task definitions, each lab contributes a distinct distribution of environments, objects, and human behaviors.

\egoversei is designed to capture a wide variety of demonstration-style tasks and behaviors across extremely diverse settings and objects. To ensure that the data is maximally useful for robot learning, we reduce noise as much as possible by enforcing that the hands remain visible in the scene whenever feasible and that demonstrators follow a consistent strategy for each sub-task. We additionally instruct demonstrators to be decisive in their actions, avoiding unnecessary hesitation or corrective motions, so that the resulting trajectories exhibit clear intent and well-defined temporal structure.

\subsection{Human Data Capture Detail}
\label{appendix:E}
\textbf{Project Aria Glasses.} We use the Project Aria Gen 1 glasses for egocentric human data capture in \egoversea. The Gen 1 device weighs approximately 75 g, designed to be worn comfortably for hours without significantly altering natural human behavior. The hardware integrates five cameras: one forward-facing global-shutter RGB cameras for egocentric scene capture, two side-facing monocolor camera for SLAM and hand tracking, and two inward-facing cameras for eye tracking. In \egoverse, we use the primary forward-facing RGB camera as the main visual stream for learning, providing a stable first-person view closely aligned with human manipulation intent and task execution. The device also includes a tightly synchronized IMU, enabling accurate visual–inertial sensing. All raw sensor streams are processed through Meta’s MPS (Machine Perception Services) pipeline, which performs calibration, temporal alignment, and visual–inertial odometry to produce metrically consistent camera poses and synchronized RGB streams in a world-referenced frame. 

\begin{figure}[h!]
\centering
\includegraphics[width=\linewidth]{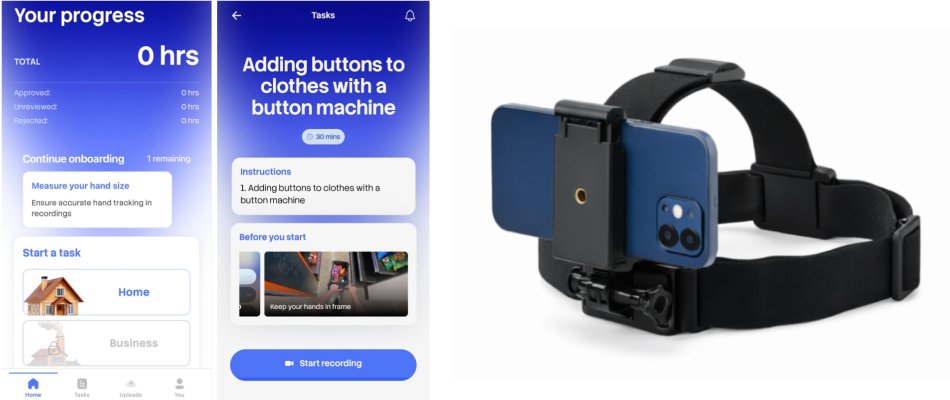}
\caption{\textbf{Phone-based Data Collection System.} (Left) Screenshot of the accompanying app for the iPhone-based human data collection system. (Right) The setup consist of an off-the-shelf head strap phone mount and an iPhone.}
\label{fig:mecka}
\end{figure}

\textbf{Phone-based Human Data Capture System.} To broaden access to large-scale human data collection beyond specialized wearable devices, we introduce a phone-based egocentric capture system built on commodity smartphones (Fig.~\ref{fig:mecka}). The setup mounts a smartphone on a lightweight head strap and records egocentric RGB video using the ultrawide camera at 1080p and 30 FPS. This configuration preserves a wide field of view over the workspace while remaining inexpensive, easy to deploy, and comfortable for extended use. Recorded videos are uploaded to a cloud processing system that estimates 6-DoF head motion via visual tracking and recovers 3D hand pose with 21 keypoints per hand. The resulting signals are temporally synchronized and converted into the same canonical representation used throughout \egoverse, including egocentric video, camera motion, and hand trajectories. By matching the output format of more instrumented capture systems, this phone-based pipeline allows data from resource-constrained contributors to be directly integrated into the broader dataset without special handling or separate learning recipe.

\textbf{Custom Capture Hardware.} \egoverse is meant to be compatible with a variety of human data sources, including highly customized capture systems designed for large-scale industrial data collection. As a concrete example, one industry partner contributes data collected using a custom stereoscopic head-mounted rig built from a pair of fisheye RGB cameras configured in a coplanar, collinear stereo setup with a fixed 6 cm baseline. The system records synchronized RGB video at 1920×1200 resolution and 30 FPS, paired with depth streams derived from stereo reconstruction and tightly time-aligned inertial measurements from an onboard IMU. Multi-sensor SLAM integrates RGB, depth, and inertial signals recovers metrically consistent 6-DoF head pose trajectories. Hand pose is estimated using vision-based models enriched with stereo depth to recover anatomically plausible 3D hand motion, from which 21 keypoints per hand are extracted and temporally smoothed. Although this rig is substantially more instrumented than the phone-based setups, all outputs are converted into the same \egoverse canonical format, allowing data collected at industrial scale to be seamlessly unified with other sources for downstream robot learning.

\subsection{\egoversedb Detail}
\label{appendix:F}
EgoDB comprises of four main components, the data collection and uploading, the SQL database, S3 based automated processing and EgoVerseDataset for training. 

\subsubsection{Data Collection and Uploading}
\egoversea data collected using the Aria glasses in the form of \emph{.vrs} files and robot data in lab-specific formats are uploaded using a unified uploading script. At upload time, data collectors are asked to annotate \texttt{operator}, \texttt{lab}, \texttt{task}, \texttt{embodiment}, \texttt{robot\_name}, \texttt{scene}, \texttt{objects}, and \texttt{is\_eval} according to the schema summarized in Table~\ref{tab:egodb_schema}. The raw files are hashed using UTC timestamps and uploaded to the S3 bucket alongside \emph{.json} files containing the annotated metadata. An hourly daemon running on a lightweight \texttt{t3.xlarge} AWS EC2 instance checks the bucket for new files and metadata and updates the SQL database accordingly. 

\subsubsection{SQL Database}
The SQL database is a Postgres SQL table with rows that correspond to the schema in Table~\ref{tab:egodb_schema} and enables easy filtering. Each row of the SQL table is a single file.

\begin{table}[t]
\centering
\small
\begin{tabular}{p{0.4\linewidth} p{0.48\linewidth}}
\toprule
\textbf{Field} & \textbf{Description} \\
\midrule
\texttt{episode\_hash} & Unique identifier for the episode, derived from UTC timestamps at upload time. \\
\texttt{operator} & Identifier for the human operator or demonstrator. \\
\texttt{lab} & Data collection site or partner lab. \\
\texttt{task} & Canonical task name (e.g., \emph{bag-grocery}, \emph{fold-clothes}). \\
\texttt{embodiment} & Embodiment type (human, robot platform, etc.). \\
\texttt{robot\_name} & Robot platform identifier, if applicable. \\
\emph{\texttt{num\_frames}} & Number of frames in the processed trajectory (updated post-processing). \\
\texttt{task\_description} & Free-form natural language description of the task. \\
\texttt{scene} & Scene or environment identifier. \\
\texttt{objects} & Serialized list of objects involved in the episode. \\
\emph{\texttt{processed\_path}} & S3 path to processed data artifacts (updated post-processing). \\
\emph{\texttt{processing\_error}} & Error message logged during automated processing, if any. \\
\emph{\texttt{mp4\_path}} & S3 path to rendered visualization video, if available. \\
\texttt{is\_deleted} & Flag indicating whether the episode has been removed or deprecated. \\
\texttt{is\_eval} & Flag indicating whether the episode is from policy evaluation. \\
\texttt{eval\_score} & Scalar evaluation score, if applicable. \\
\texttt{eval\_success} & Binary indicator of task success during evaluation. \\
\bottomrule
\end{tabular}
\caption{Schema of the EgoDB episode table. Fields in italics are updateable during automated processing.}
\label{tab:egodb_schema}
\end{table}

\subsubsection{Ray Processing Daemon}
\egoversea and Robot data are processed and have their metadata updated by nightly Ray processing daemons. The daemon consists of 3 Ray clusters.

\noindent\textbf{Project Aria Data. }For the \egoversea data, Cluster A responsible for running MPS (Machine Perception Services) runs on a single head node (\emph{t3a.2xlarge}). It syncs batches of files without corresponding MPS folders and makes nightly parallelized API calls to the MPS platform. After MPS is completed, it syncs back the processed MPS folders containing the data to the S3 bucket. Cluster B is responsible for converting the MPS output into the training ready processed format. This cluster has a \emph{t3a.2xlarge} head node and \emph{r6a.2xlarge} worker nodes. The head node checks the SQL table for the entries without a \texttt{processed\_path} and with corresponding MPS folders and schedules jobs on worker nodes to be processed.

\noindent{\textbf{Robot Data. }}For the robot data, Cluster C is responsible for converting the raw robot data into the training ready processed format. This cluster has a \emph{t3a.2xlarge} and \emph{c5.18xlarge} worker nodes. The head node checks the SQL table for the entries without a \texttt{processed\_path} and with corresponding MPS folders and schedules jobs on worker nodes to be processed.

The worker nodes are additionally responsible for producing an mp4 of the processed data, and return a processed path, an error if applicable and number of frames in the file to the head node. Once the worker nodes process the files, the corresponding head nodes update the following sections of the SQL table, \texttt{processed\_path}, \texttt{mp4\_path}, \texttt{processing\_error} and \texttt{num\_frames} for those files.

\subsubsection{EgoVerseDataset}
We provide a unified dataset interface, \texttt{EgoVerseDataset}, for loading EgoVerse data directly from S3 into training-ready PyTorch datasets. The dataset enables scalable, filtered access to large collections of processed episodes across embodiments, tasks, and labs.

\texttt{EgoVerseDataset} resolves valid episodes by querying the SQL database using user-specified filters (e.g., task, lab, scene, robot name) and retrieves only episodes with a populated \texttt{processed\_path}. Matching episodes are synchronized from S3 to a local cache directory, with safeguards to avoid re-downloading episodes that are already present locally. The dataset downloading is parallelized using the \emph{s5cmd} library which is a Python parallelized syncing tool for AWS S3. Each episode is loaded as an independent dataset instance, and dataset construction is parallelized across episodes to reduce load time. Episodes whose embodiment metadata does not match the requested embodiment are automatically skipped. The resulting collection of episodes is split into training and validation subsets according to a configurable validation ratio, or alternatively subsampled using a percent-based split.

The dataset supports multiple operating modes, including \texttt{train}, \texttt{valid}, \texttt{total}, and \texttt{percent}, enabling consistent dataset construction for both standard training and scaling experiments. All loaded episodes are exposed through a unified interface, allowing downstream training code to treat the aggregated data as a single iterable source.
\subsubsection{Accessing Data from S3.}
Given a set of metadata filters, data are resolved from the SQL database, synchronized from S3, and instantiated as PyTorch dataset objects. Code block ~\ref{lst:s3_access_code} shows a simplified example illustrating this process.







\begin{lstlisting}[language=Python, caption={Simplified example illustrating SQL-based episode resolution, S3 synchronization using \texttt{s5cmd}, and instantiation of training datasets.}, label={lst:s3_access_code}]
# 1. Query SQL table to resolve processed episodes
filters = {
    "robot_name": "robot_a",
    "lab": "lab_a",
    "task": "task_x",
    "is_deleted": False,
}
rows = query_sql_table(filters)
# rows: [(processed_path, episode_hash), ...]

# 2. Download processed data from S3
for processed_path, episode_hash in rows:
    local_dir = f"/tmp/egoverse/{episode_hash}/"
    s3_src = f"{processed_path}/*"
    run_command(["s5cmd", "sync", s3_src, local_dir])

# 3. Instantiate dataset objects
datasets = []
for _, episode_hash in rows:
    dataset = SingleEgoVerseDataset(
        root=f"/tmp/egoverse/{episode_hash}",
        mode="train",
    )
    datasets.append(dataset)

# 4. Combine datasets for training
train_dataset = MultiEgoVerseDataset(datasets)
\end{lstlisting}

\subsection{Data Alignment and Post Processing}
\label{appendix:G}
Due to inherent differences in execution speed between humans and robot teleoperation, we apply post-processing to temporally align action trajectories across embodiments. For human demonstrations, we extract a $1$-second window of actions and resample it to a sequence of length $T=100$. For robot data, we extract a $1.5$-second window and similarly resample it to length $100$.
This ensures that all trajectories represent a comparable phase of task execution despite differences in control frequency and execution speed. We apply linear interpolation for the 3D positions in the actions and spherical linear interpolation (SLERP) for quaternion and euler angle rotation representations.

\subsection{Robot Data Collection Setup Detail}
\label{appendix:H}
\subsubsection{Hardware Setup}
\paragraph{\robotGT} We employ a VR teleoperation system using the Meta Oculus 3 headset and Oculus Pro controllers based on the RAIL Lab Oculus Reader~\cite{OrbikEbert2021OculusReader}. We use inverse kinematics on the commanded robot base frame end-effector using the Mink IK Solver~\cite{Zakka_Mink_Python_inverse_2025} to obtain joint angles. The joint angles are executed by the ARX5 Joint Space Controller. The system is implemented using a multi-threaded Python setup, while the low-level hardware communication is handled via a CAN-based interface. The robot platform consists of two off-the-shelf ARX5 robot arms mounted as shown in Fig.~\ref{fig:robots} using off-the-shelf Vention aluminum beams with 3D-printed connectors.

\paragraph{\robotStanford} We employ a customized 3D-printed GELLO \cite{wu2023gello} device for teleoperation, with a one-to-one joint mapping between the GELLO interface and the ARX robot. The robot platform consists of two off-the-shelf ARX5 robot arms mounted left and right on a rigid 4040 aluminum frame with 3D-printed connectors as shown in Fig.~\ref{fig:robots}. The gripper opening width is controlled via a trigger mounted at the distal end of each GELLO arm. The system is implemented using ROS 2 for message passing and inter-module communication. Low-level hardware communication is handled via a CAN-based interface, while motion execution is managed by the standard ARX5 Cartesian Controller.

\paragraph{\robotUCSD} We use a Unitree G1 robot equipped with Inspire 6dof hands for experiments. The robot is equipped with a 3 DoF actuated head~\cite{li2025amoadaptivemotionoptimization}. An Apple Vision Pro~\cite{cheng2024tv} for teleoperation, which maps the left and right wrist poses to the robot left wrist and right wrist, and use fingertip keypoints for retargeting. The head motors track the rotation fo the teleoperator’s head via IK.

The controller runs on the Jetson Orin NX which is built in to the G1 robot. We implemented a python interface that communicates with the Unitree C++ backend via LCM. For inference, we deploy the model on a remote server, and use websocket for synchronized states and images communication.

\subsubsection{Robot Data Composition}
The per-task amount of robot demonstrations per robot and per task is summarized in Table~\ref{tab:task_demos}.
\begin{table}[t]
\centering
\begin{tabular}{lccc}
\toprule
\textbf{Task} & \multicolumn{3}{c}{\textbf{\# Demos \,|\, \# Hours}} \\
\cmidrule(lr){2-4}
 & \robotGT & \robotStanford & \robotUCSD \\
\midrule
object-in-container & 100 \,|\, 1.2 & 200 \,|\, 2.7 & 240 \,|\, 3.0 \\
bag-grocery       & 300 \,|\, 5.1 & 150 \,|\, 1.67 & 139 \,|\, 1.8 \\
cup-on-saucer       & 360 \,|\, 3.3 & 183 \,|\, 1.0 & 111 \,|\, 1.2 \\
fold-clothes        & 300 \,|\, 3.0 & -- & -- \\
\bottomrule
\end{tabular}
\caption{Robot dataset composition across tasks and platforms, reported as number of demonstrations and total hours.}
\label{tab:task_demos}
\end{table}

\subsection{Policy Architecture and Learning Detail}
\label{appendix:I}

All learning hyperparameters are specified in Table~\ref{tab:hyperparams}
\subsubsection{Cross Embodiment Encoder and Stems}
\paragraph{\textbf{Vision Inputs}} Given an RGB frame \(\mathbf{I}\!\in\!\mathbb{R}^{H\times W\times 3}\), we apply ImageNet normalization an d then pass it through a ResNet-18 encoder truncated \emph{before} the global-pool layer to obtain a \(7\times7\times512\) feature map. We flatten the last convolutions and project them to $d_{proj}$ with a single linear layer. We cross-attend $L$ learnable query tokens of dimension $d$ to the projected features using a single multi-head cross attention block with $D_{stem}$ heads. 

\paragraph{\textbf{Proprioceptive Inputs}} The proprioceptive observation vector \(\mathbf{q}\!\in\!\mathbb{R}^{d_{q}}\) (joint angles, end effector pose etc.) is quantile normalized and passed through a single linear layer of hidden dimension $d_{proj}$. A single multi-head cross attention block with $D_{stem}$ heads attends \(L\) query tokens with hidden dimension $d$ to the projected proprioceptive data. 

\paragraph{\textbf{Encoder}} Input observation data from $m$ stems produces $m \times L$ query tokens. These tokens are concatenated along the sequence dimension. A set of learnable context tokens $M$ are prepended to the token sequence. The new sequence of $M + m \cdot L$ is the input to the cross embodiment encoder. The cross embodiment encoder $f_{\phi}$ is a multi-block transformer encoder without masking. It consists of $D_{enc}$ heads, $N_{enc}$ blocks, and an embedding dimension of $d$. Each head has an attention dimension of $d / D_{enc}$. 

\subsubsection{Flow Matching Decoder.}
The decoder is parameterized by a multi-block diffusion transformer with
$N_{\mathrm{dec}}$ layers, $D_{\mathrm{dec}}$ attention heads, and embedding
dimension $d_{\mathrm{dec}}$. The $M$ context tokens produced by the encoder are
used as the conditioning sequence for the flow-matching decoder.

A noise token sequence of shape $\mathbb{R}^{T \times d_{\mathrm{dec}}/2}$ is
combined with a learnable positional embedding. A sine--cosine embedding of the continuous time variable
$\tau \sim \mathrm{Beta}(1.5, 1.0)$ is expanded to
$\mathbb{R}^{T \times d_{\mathrm{dec}}/2}$ and concatenated along the hidden
dimension, yielding a token sequence of dimension
$\mathbb{R}^{T \times d_{\mathrm{dec}}}$. During training, the noise token sequence is constructed by sampling i.i.d.
Gaussian noise $a_0 \sim \mathcal{N}(0, I)$ with shape
$\mathbb{R}^{T \times d_a}$ and sampling a continuous timestep
$\tau \in (0,1]$. The decoder input is formed by linear interpolation between
noise and the ground-truth action sequence $a_1 = \mathbf{a}$,
\[
x_\tau = \tau a_0 + (1 - \tau) a_1,
\]
which is then projected into the decoder token space.

At inference time, the noise token sequence is initialized as pure Gaussian
noise $x_{\tau=1} \sim \mathcal{N}(0, I)$. The decoder is applied iteratively
while integrating the learned velocity field from $\tau=1$ to $\tau=0$ using
fixed-step Euler updates, producing the final action sequence at the end of
integration. 10 fixed step Euler updates (inference steps) are used during evaluation.

The resulting token sequence is processed by a stack of transformer blocks with
alternating self-attention and cross-attention. In the cross-attention blocks,
the action tokens attend to the $M$ conditioning tokens from the encoder. A final
linear projection maps the hidden tokens to the predicted action sequence in
$\mathbb{R}^{T \times d_a}$.

\noindent\subsubsection{Co-training with Flow Matching.}
As discussed earlier, the total co-training loss is defined as
\[
\mathcal{L}_{\text{BC-cotrain}}
=
\mathcal{L}_{\text{CFM}}^{\text{robot}}
+
\mathcal{L}_{\text{CFM}}^{\text{human}}.
\]
For a given embodiment $e$, we sample a timestep
$\tau \sim \mathrm{Beta}(1.5, 1.0)$ and minimize the error in the predicted vector
field:
\[
\mathcal{L}_{\text{CFM}}^{e}
=
\mathbb{E}_{\tau, a_0, a_1, s}
\Big[
\big\|
\pi_\theta(x_\tau, \tau, f_\phi(s))
-
(a_0 - a_1)
\big\|^2
\Big],
\]
where $x_\tau = \tau a_0 + (1 - \tau) a_1$ denotes the linear probability path
between Gaussian noise $a_0$ and the target action $a_1$.

\subsubsection{Training Details.}
We train the model for 150{,}000 optimization steps with a global batch size of
32--64 and learning rate of $1 \times 10^{-4}$. Since experiments are conducted across multiple labs and platforms, the
available compute resources and total training time vary. \

All model hyperparameters are summarized in Table~\ref{tab:hyperparams}.
\begin{table}[t]
\centering
\small
\begin{tabular}{l p{0.40\linewidth}}
\toprule
\textbf{Hyperparameter} & \textbf{Value} \\
\midrule
\multicolumn{2}{l}{\emph{Observation Stems}} \\
Visual Feature Embeddings & ResNet-18 \\
$m$ (number of stems) & 4 (1 vision main camera, 2 vision wrist camera + 1 proprioception) \\
$L$ (query tokens per stem) & 16 \\
$D_{\text{stem}}$ & 8 \\
$d_{\text{proj}}$ & 256 \\
\midrule
\multicolumn{2}{l}{\emph{Cross-Embodiment Encoder}} \\
$M$ (context tokens) & 64 \\
$N_{\text{enc}}$ & 16 \\
$D_{\text{enc}}$ & 8 \\
$d$ & 256 \\
Encoder positional embedding & sine-cosine \\
Encoder normalization & Pre-LN \\
\midrule
\multicolumn{2}{l}{\emph{Flow Matching Decoder}} \\
$N_{\text{dec}}$ & 6 \\
$D_{\text{dec}}$ & 4 \\
$d_{\text{dec}}$ & 128 \\
$d_a$ & task-specific \\
Time embedding & sine--cosine \\
$\tau$ distribution & $\mathrm{Beta}(1.5, 1.0)$ \\
Probability path interpolation & linear \\
\midrule
\multicolumn{2}{l}{\emph{Training}} \\
Optimizer & AdamW \\
Learning rate & $1\times10^{-4}$ \\
Weight decay & $1\times10^{-4}$ \\
Training steps & 150{,}000 \\
Global batch size per embodiment & 32--64 \\
Human : robot batch ratio & 1 : 1 \\
\midrule
\multicolumn{2}{l}{\emph{Inference}} \\
ODE solver & Euler \\
Integration steps & 10 \\
Integration interval & $\tau : 1 \rightarrow 0$ \\
Initial noise & $\mathcal{N}(0, I)$ \\
\bottomrule
\end{tabular}
\caption{Model hyperparameters. Symbols follow the notation introduced in Appendix~\ref{appendix:J}.}
\label{tab:hyperparams}
\end{table}


\subsection{Robot Experiment Results}
\label{appendix:J}

\subsubsection{Training Mixture Details}
We summarize training data mixtures for the various results reported in the paper below.

\textbf{Flagship Co-train (EV(8hr) + ID(2hr)):} For each task, we use a fixed co-training setup that combines 8 hours of \egoversea (EV) human data with 2 hours of in-domain (ID) human data, together with task-matched robot demonstrations. The in-domain human data are collected using the same objects, task configuration, and scene as the robot dataset for the particular task. The \egoversea data are aggregated across four collection sites for that task. Human samples are uniformly drawn from the union of EV (8hr) and ID (2hr) data. Robot data are sampled uniformly from the available demonstrations for the task and platform, as summarized in Table~\ref{tab:task_demos}, using frame-level sampling. 

\textbf{Scaling Law Experiments:} This experiment is to ablate the effect of each human data component: In Domain Human (ID) and \egoversea Human (EV). The main result is discussed in Sec.~\ref{sec:experiment:realworld}.  Here we explain the data composition for each training setup.
\begin{itemize}
    \item \emph{EV(8hr)}: Flagship Co-train training mixture without the in-domain human data.
    \item \emph{EV(2hr) + ID(2hr)}: Flagship Co-train model training mixture with only 1 site's \egoversea data. 
    \item \emph{ID(2hr)}: Flagship Co-train model training mixture without the EV(8hr) data.
    \item \emph{ID(1hr)}: Flagship Co-train model training mixture without the EV(8hr) data and in-domain data with demonstrations subsampled to 1 hour.
    \item \emph{Robot only}: Flagship Co-train model training mixture without any human data.
\end{itemize}

For both the Flagship Co-training and the Scaling Law experiments, we use a global batch size of 32 with a fixed 1:1 human-to-robot ratio, i.e., 16 human samples and 16 robot samples per batch. This co-training recipe is used for all flagship experiments; variations are described separately in the scaling-law experiments.


\subsubsection{Rollout Evaluation Protocol} 
\begin{figure}[t]
\centering
\includegraphics[width=\columnwidth]{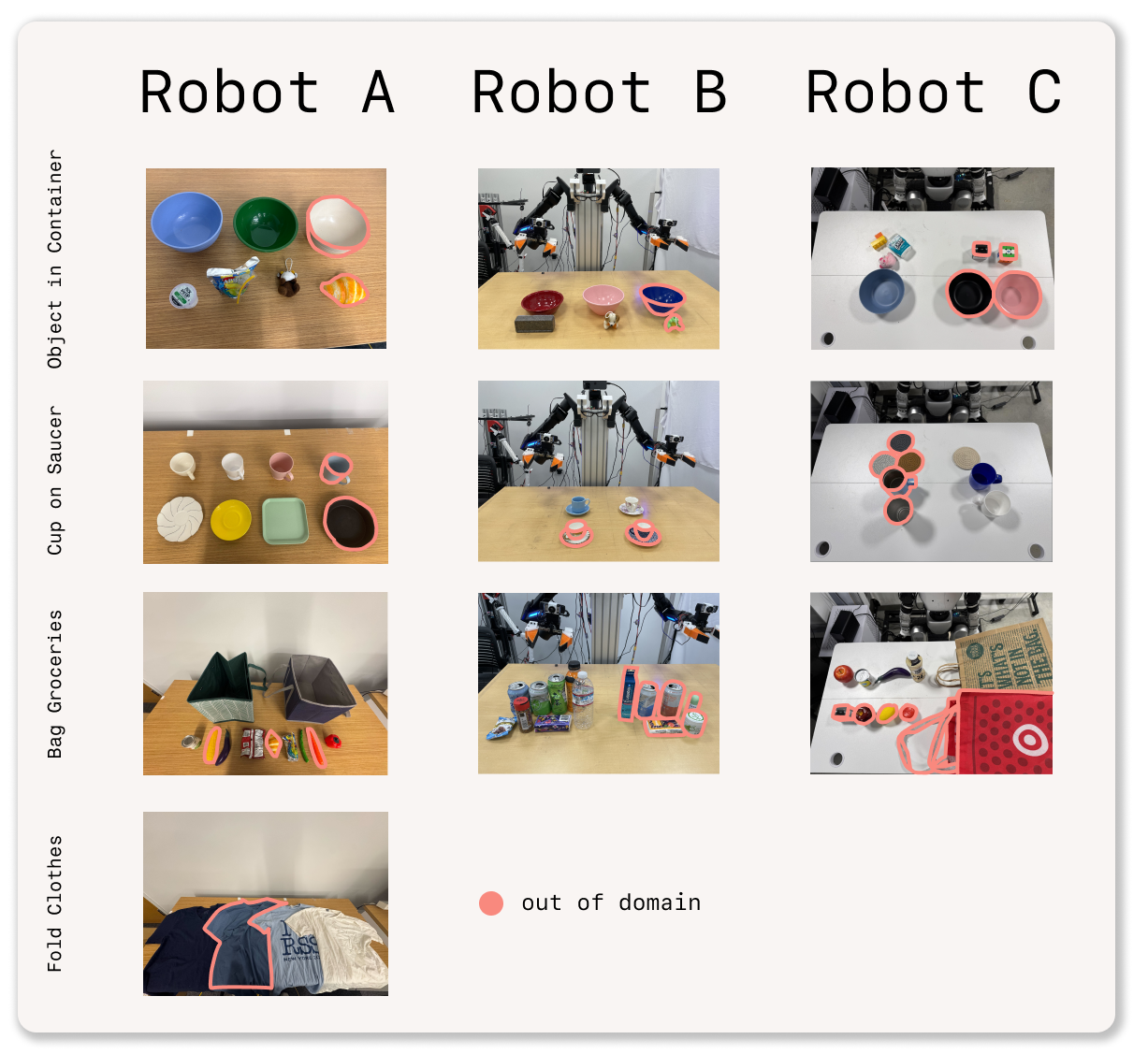}
\caption{Objects used for training and evaluation across tasks and robot platforms.}
\label{fig:obs}
\end{figure}

We provide more detail of standardized evaluation protocol shared across different labs (robots) below. Images of the training and evaluation objects are in Fig~\ref{fig:obs}.

\paragraph{\textbf{object-in-container}} The scene contains one object and one container randomly place with the object being closer to the robot vertically than the container.

\emph{In-domain}: Randomly sample 20 positions across the workspace grid for object and container. For the first 10 positions, use 1 object-container combination from training set and for the next 10 use another object-container combination. Perform one rollout for each position. 

\emph{Out-of-domain}: Sample an object-container combination from the human dataset that is not present in the robot dataset. For the first 10 rollouts, randomly sample 10 positions on the in-domain table. For the second 10 rollouts, randomly sample 10 positions on an unseen table. Perform one rollout for each position. Run each rollout for 40 seconds. 

\emph{Termination: } Rollouts are terminated early if the policy becomes stuck, exhibits unsafe behavior, or is unable to continue execution.

\emph{Scoring: } Score the policy 1pt for each successful object placement in container and 1pt for each successful emptying of the container. Report total points obtained within 40 seconds. 

\paragraph{\textbf{cup-on-saucer}} 
The scene starts with a cup placed to either the right or left of the workspace and a saucer placed on the opposite side.

\emph{In-domain}: For each case (cup on left and cup on right), randomly sample 10 positions across the workspace grid, yielding 20 total rollouts. At each position, randomly select a cup--saucer combination from the seen object set and apply a random planar rotation sampled uniformly from $\pm 30^{\circ}$. Perform one rollout per position.

\emph{Out-of-domain}: For each case (cup on left and cup on right), randomly sample 5 positions on the in-domain table and 5 positions on an unseen table, yielding 10 total rollouts per case. At each position, randomly select a cup--saucer combination from the seen object set and apply a random planar rotation sampled uniformly from $\pm 30^{\circ}$. Perform one rollout per position.

\emph{Termination}: Rollouts are terminated early if the policy becomes stuck, exhibits unsafe behavior, or is unable to continue execution.

\emph{Scoring}: Assign 1 point for successfully rotating and picking up the cup, 1 point for a successful handover, and 1 point for placing the cup on the saucer. Additionally, report the total task success rate (SR), defined as the percentage of rollouts in which the cup is correctly placed on the saucer.

\paragraph{\textbf{bag-grocery}} The scene contains a bag and three grocery objects. The bag is placed to the left of the robot, and the three objects are placed to the right.

\emph{In-domain}: Randomly sample 20 positions across the workspace grid for the bag and the three objects. For each position, sample a set of 3 objects from the training set and a random permutation (ordering) of those objects. Perform one rollout per position.

\emph{Out-of-domain}: Randomly sample 10 positions across the workspace grid. For the first 5 positions, sample one random 3-object set; for the next 5 positions, sample a second random 3-object set. For each position, randomize the permutation (ordering) of the objects. Repeat the same protocol on an unseen scene.

\emph{Termination}: Rollouts are terminated early if the policy becomes stuck, exhibits unsafe behavior, or is unable to continue execution.

\emph{Scoring}: Assign 1 point for successfully opening the bag and 1 point for each object placed in the bag (3 points), for a maximum of 4 points per rollout. A placement is considered successful only if the objects are placed in left-to-right order inside the bag.

\paragraph{\textbf{fold-clothes}} The scene contains a t-shirt placed horizontally, with the collar oriented to the right of the robot.

\emph{In-domain}: Randomly select 2 t-shirts from the training set. Randomly sample 10 positions across the workspace grid and apply a random planar rotation from $\{0^{\circ}, 30^{\circ}, -30^{\circ}\}$. For each sampled position, perform one rollout for each of the 2 selected t-shirts (2 rollouts per position).

\emph{Out-of-domain}: Randomly select 1 t-shirt that appears only in the human dataset (and is not present in the robot dataset). Randomly sample 10 positions across the workspace grid and apply a random planar rotation from $\{0^{\circ}, 30^{\circ}, -30^{\circ}\}$. Perform one rollout per position. Repeat the same protocol on an unseen scene.

\emph{Termination}: Rollouts are terminated early if the policy becomes stuck, exhibits unsafe behavior, or is unable to continue execution.

\emph{Scoring}: Assign 1 point for successfully completing the bottom-sleeves folding stage, 1 point for successfully completing the top-sleeves folding stage, and 1 point for folding the t-shirt in half, for a maximum of 3 points per rollout. Full completion of all three stages denotes task success; report task success rate (SR) as the percentage of rollouts that achieve full completion.

\emph{Score Normalization} 
We compute normalized score for \emph{bag-grocery} and \emph{object-in-container} to standardize comparisons across different execution speeds across labs. We report success rates for fold clothes and cup on saucer.  We calculate normalized score as total points scored divided by maximum possible points.

\subsubsection{Additional Analysis}
\begin{figure}[t]
\centering
\includegraphics[width=\columnwidth]{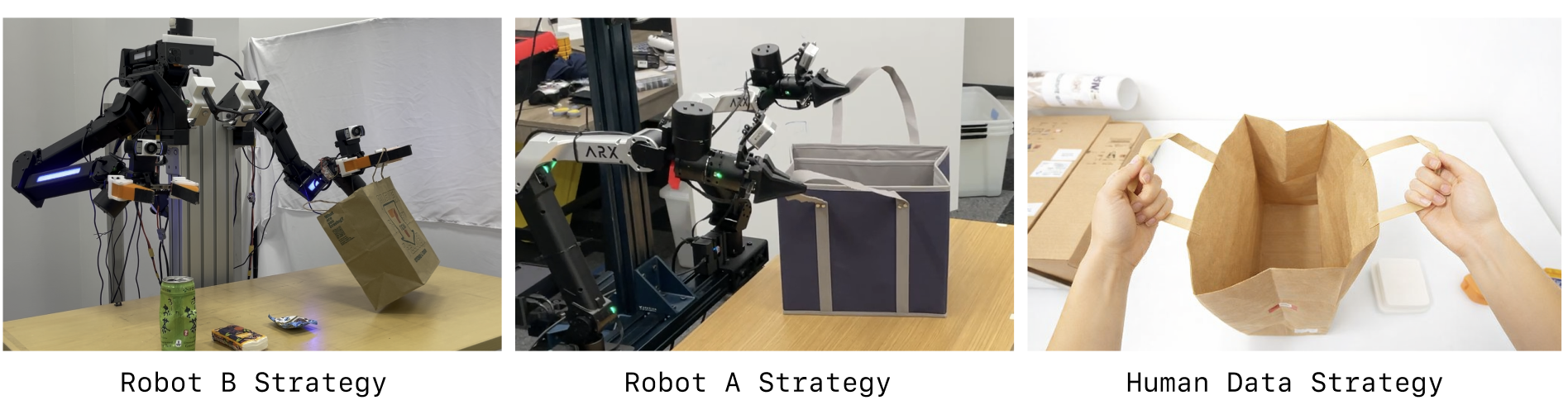}
\caption{Differences in task execution strategies for the \textit{bag-grocery} task across embodiments.}
\label{fig:strategy}
\end{figure}

As discussed in Sec.~\ref{sec:experiment:realworld}, \robotStanford exhibits a systematic strategy mismatch between human and robot demonstrations, which we hypothesize contributes to the observed degradation in co-training performance. This mismatch is illustrated in Fig.~\ref{fig:strategy}. In both the \egoversea human demonstrations and the \robotGT robot demonstrations, the bag is first opened using two hands (or grippers), after which grocery items are inserted. In contrast, the \robotStanford demonstrations employ a different strategy, where one gripper is used to prop the bag open while the other performs item insertion. This divergence leads to inconsistent behavior distributions between human and robot data, potentially weakening cross-embodiment alignment during co-training.

\begin{figure}[t]
\centering
\includegraphics[width=\columnwidth]{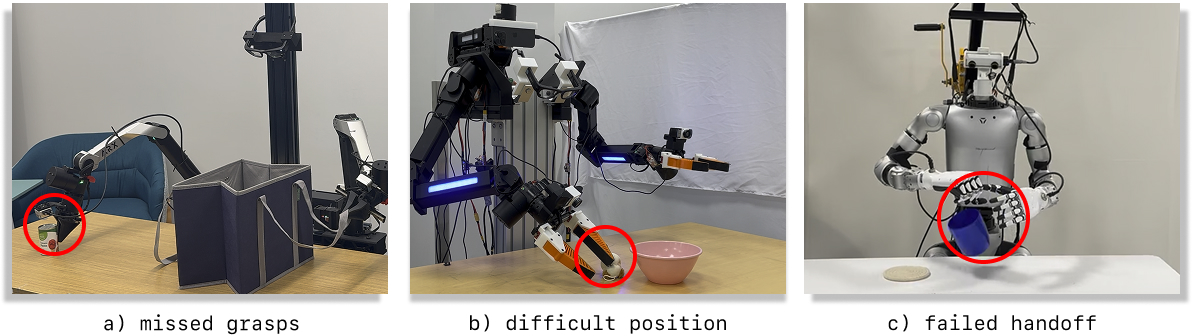}
\caption{Common failure modes for each task}
\label{fig:failure}
\end{figure}

\subsubsection{Task Failure Modes}
For \emph{object-in-container} and \emph{bag-grocery}, we saw difficulty with picking primitives in certain parts of the workspace. While our co-trained policies generally exhibited more robust grasping primitives, there is still room for improvement.  For the \emph{cup-on-saucer} task, the object handover was difficult, especially for Robot C with a dexterous hand. Common failure modes are shown in Fig~\ref{fig:failure}.

\subsection{Controlled Diversity Experiments}
\label{appendix:K}

We provide detailed experimental protocols and data compositions for the controlled diversity studies. While Sec.~\ref{sec:experiment:diversity} presents results on \textit{fold-clothes}, this section presents the corresponding results for \textit{cup-on-saucer} and a precise training-validation setting used across all four scaling regimes. 

\noindent \textbf{Data Composition and Evaluation Protocols.} Experiments are conducted using the controlled-diversity subset described in Sec.~\ref{ssec:dataset:protocol}, consisting of 16 demonstrators and 16 scenes for each task. 
Following the same policy architecture and training procedure described in Sec.~\ref{appendix:J}, we vary the training data by constructing different demonstrator–scene pair combinations and evaluate human open-loop action prediction accuracy using the offline Avg-MSE metric: given a predicted action sequence $\hat{\mathbf{a}}_{1:T}\in\mathbb{R}^{T\times D}$ and ground-truth sequence $\mathbf{a}_{1:T}\in\mathbb{R}^{T\times D}$, we compute the mean-squared error at each timestep and average over the sequence and action dimensions:
    \[
        \mathrm{Avg\text{-}MSE}(\hat{\mathbf{a}}_{1:T}, \mathbf{a}_{1:T})
        = \frac{1}{T}\sum_{t=1}^{T}\frac{1}{D}\left\lVert \hat{\mathbf{a}}_{t}-\mathbf{a}_{t}\right\rVert_2^2,
    \]
where we report this value averaged across validation episodes. 
We provide the detailed training data budgets for single-scene demonstrator scaling, multi-scene demonstrator scaling, mixed diversity scaling, and scene scaling in Table~\ref{tab:single_scene_demo_scaling_budget}, Table~\ref{tab:multi_scene_demo_scaling}, Table~\ref{tab:mixed_diversity_scaling}, and Table~\ref{tab:scene_diversity_budget}, respectively.

\begin{figure}[t]
    \centering
    \includegraphics[width=\linewidth]{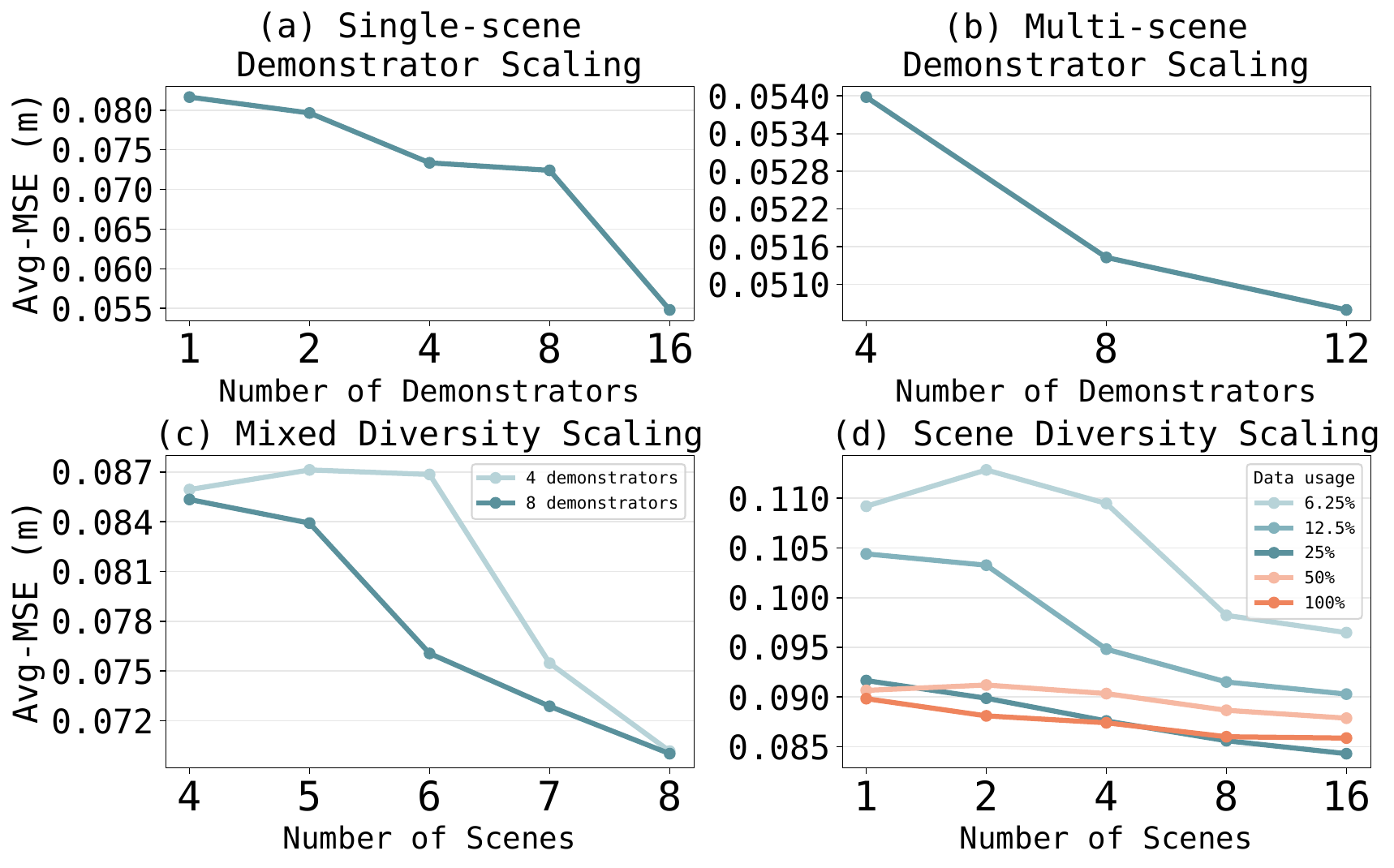}
    \caption{\textbf{Controlled diversity results on fold-clothes.}}
    \label{fig:appendix_fold_clothes_diversity}
\end{figure}

\begin{figure}[t]
    \centering
    \includegraphics[width=\linewidth]{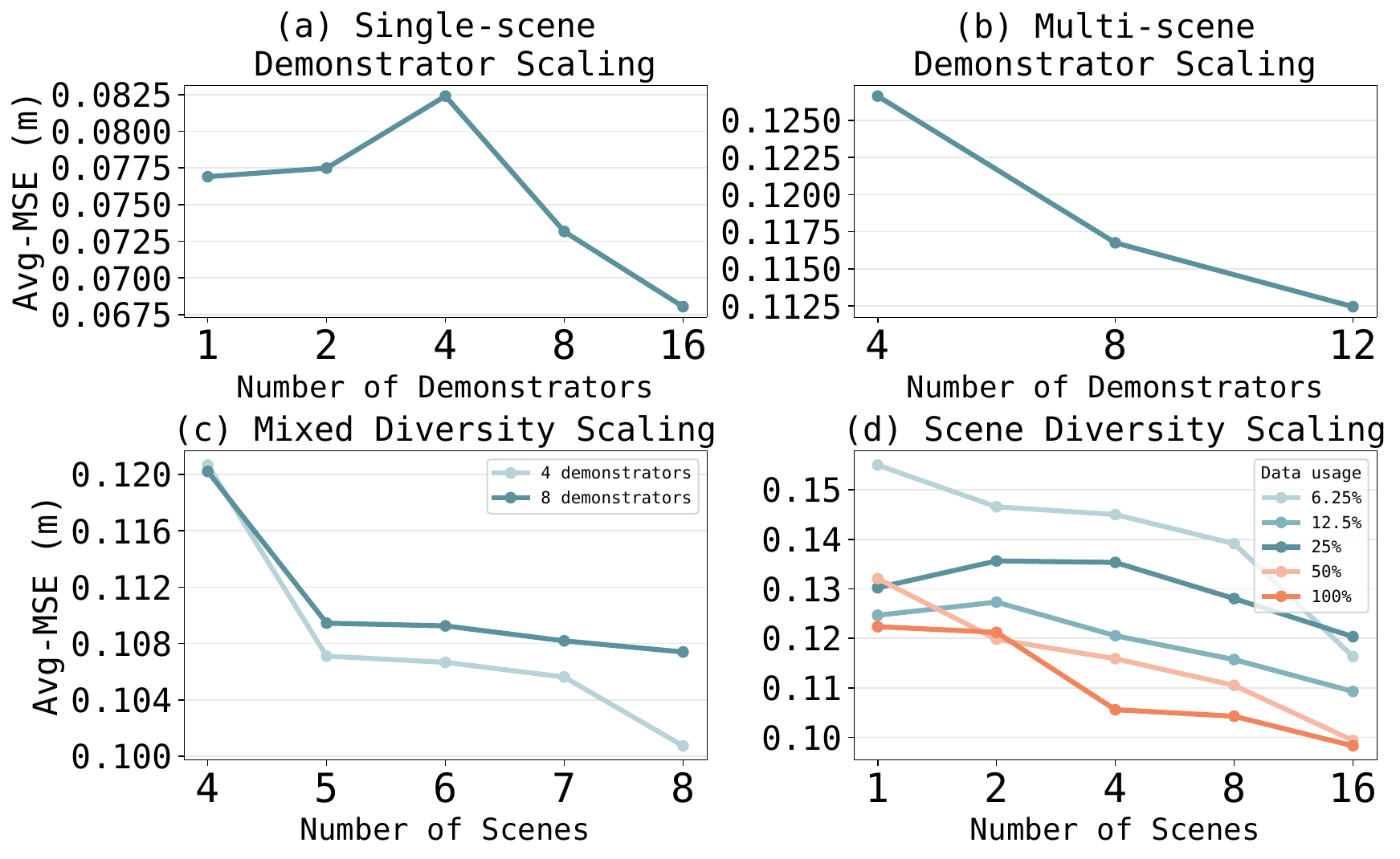}
    \caption{\textbf{Controlled diversity results on cup-on-saucer.}}
    \label{fig:appendix_cup_on_saucer_diversity}
\end{figure}

\noindent \textbf{Analysis of Controlled Diversity Experiments.} We present the results for \textit{fold-clothes} and \textit{cup-on-saucer} in Fig.~\ref{fig:appendix_fold_clothes_diversity} and Fig.~\ref{fig:appendix_cup_on_saucer_diversity}, respectively.

\subsubsection{Single-scene Demonstrator Scaling} This experiment studies whether adding motion diversity from increasing the number of demonstrators, given a fixed data budget of 2 hours, improves generalization towards unseen demonstrator at the same scene. As shown in Figs.~\ref{fig:appendix_fold_clothes_diversity}(a),~\ref{fig:appendix_cup_on_saucer_diversity}(a), increasing the number of demonstrators improves performance across both tasks. For \textit{fold-clothes}, Avg-MSE decreases monotonically with demonstrator count, with larger gains at higher diversity levels, indicating that demonstrator variation meaningfully enhances coverage of state-action distribution. For \textit{cup-on-saucer}, performance is slightly non-monotonic at low counts but improves at larger scales, suggesting that sufficient demonstrator diversity is needed to yield stable gains.

\subsubsection{Multi-scene Demonstrator Scaling} In this study, we extend demonstrator scaling from a fixed single scene to eight scenes to examine whether the scaling effect persists in a multi-scene setting, aligning more closely with our real-world data collection setup. It is evaluated on unseen demonstrators within the same scenes.
Given a fixed training data budget of 8 hours, the multi-scene demonstrator scaling results (Figs.~\ref{fig:appendix_fold_clothes_diversity}(b),~\ref{fig:appendix_cup_on_saucer_diversity}(b)) show consistent and clear improvements in generalization to unseen demonstrators as the number of demonstrators increases from 4 to 12 across both tasks. 

\subsubsection{Scene Diversity Scaling} Next, we assess how scene diversity and per-scene data allocation affect scene generalization, evaluated on unseen scenes data collected from other labs. In both tasks, increasing the number of scenes consistently reduces Avg-MSE, demonstrating that scene diversity improves generalization, as shown in Figs.~\ref{fig:appendix_fold_clothes_diversity}(d),~\ref{fig:appendix_cup_on_saucer_diversity}(d). 
In \textit{fold-clothes}, the trend is monotonic and especially pronounced under low data budgets, indicating that broader environmental coverage effectively compensates for limited per-scene data, while per-scene data quantity becomes less critical once the overall budget is sufficiently large.
In \textit{cup-on-saucer}, the decrease is slightly less smooth but still consistent overall, with stronger gains under higher data usage, suggesting that this task benefits from both sufficient per-scene data and expanded scene diversity.

\subsubsection{Mixed Diversity Scaling} Under a fixed 4-hour data budget, we study the joint effect of scaling scene diversity (from 4 to 8 scenes) and demonstrator diversity (from 4 to 8 demonstrators), evaluating on unseen demonstrators and scenes collected in other labs. As shown in Figs.~\ref{fig:appendix_fold_clothes_diversity}(c) and~\ref{fig:appendix_cup_on_saucer_diversity}(c), increasing scene diversity consistently reduces Avg-MSE for both \textit{fold-clothes} and \textit{cup-on-saucer}, indicating that broader environmental coverage is a reliable driver of scene generalization. 
Demonstrator diversity provides an additional but task-dependent benefit: for \textit{fold-clothes}, using 8 demonstrators consistently outperforms 4, suggesting that increased demonstrator variation yields a richer motion distribution that better covers unseen scenarios. 
In contrast, for \textit{cup-on-saucer}, the 4-demonstrator setting slightly outperforms 8 demonstrators, implying that for this more constrained task, additional demonstrator-induced behavioral noise may outweigh its coverage benefits, making generalization primarily driven by scene variation.

\noindent \textbf{Summary.} We conduct controlled diversity studies on \textit{fold-clothes} and \textit{cup-on-saucer} using a standardized dataset of 16 demonstrators and 16 scenes, systematically varying demonstrator and scene diversity under fixed data budgets and evaluating generalization with the offline Avg-MSE metric. Across four scaling regimes, we find consistent evidence that \textit{increasing diversity improves generalization}, with scene diversity serving as the most reliable driver across tasks. Demonstrator diversity provides additional gains, particularly for \textit{fold-clothes}, where demonstrator variation meaningfully enriches the motion distribution, while its benefit is more task-dependent for \textit{cup-on-saucer}. Overall, these results highlight the complementary roles of scene and demonstrator diversity in shaping generalization for human data, with their relative importance varying by task structure and constraint.

\begin{table}[t]
\centering
\caption{\textbf{Single-scene demonstrator scaling (fixed 2-hour budget).} Training data are collected in one fixed scene. As the number of training demonstrators increases, the per-demonstrator data duration decreases so that the total budget remains 2 hours. Evaluation is performed on a held-out 17th demonstrator with 7.5 mins of data in the same scene. We refer to a demonstrator–scene pair as a DS Pair. }
\label{tab:single_scene_demo_scaling_budget}
\setlength{\tabcolsep}{10pt}
\renewcommand{\arraystretch}{1.2}
\begin{tabular}{c c }
\hline
\textbf{\# Train Demonstrators} & \textbf{Mins / DS Pair} \\
\hline
1  & 120.0  \\
2  & 60.0   \\
4  & 30.0   \\
8  & 15.0   \\
16 & 7.5    \\
\hline
\end{tabular}
\end{table}

\begin{table}[t]
\centering
\caption{\textbf{Multi-scene demonstrator scaling (fixed 8-hour budget).} Training data are collected across 8 fixed scenes. As the number of training demonstrators increases, the per-demonstrator data duration decreases proportionally to maintain the total budget of 8 hours. Evaluation is conducted on unseen demonstrators within the same 8 scenes. We refer to a demonstrator–scene pair as a DS Pair.}
\label{tab:multi_scene_demo_scaling}
\setlength{\tabcolsep}{10pt}
\renewcommand{\arraystretch}{1.2}
\begin{tabular}{c c}
\hline
\textbf{\# Train Demonstrators} & \textbf{Mins / DS Pair} \\ \hline
4  & 15.0 \\
8  & 7.5  \\
12 & 3.75  \\ \hline
\end{tabular}
\end{table}

\begin{table}[h!]
\centering
\caption{\textbf{Mixed diversity scaling (fixed 4-hour budget).} This table details the distribution of the fixed 4-hour data budget across varying numbers of scenes and demonstrators. Scaling both axes allows for the study of joint environmental and motion diversity effects. Evaluation is conducted on unseen demonstrators and scenes. We denote a demonstrator as D, a scene as S, and a demonstrator–scene pair as a DS pair.}
\label{tab:mixed_diversity_scaling}
\setlength{\tabcolsep}{10pt}
\renewcommand{\arraystretch}{1.20}
\footnotesize
\begin{tabular}{c c c c c}
\hline
\textbf{\# S} & \textbf{\# D} &
\textbf{Mins / D} & \textbf{Mins / S} & \textbf{Mins / DS Pair} \\
\hline
4 & 4 & 60.0 & 60.0 & 15.00 \\
5 & 4 & 60.0 & 48.0 & 12.00 \\
6 & 4 & 60.0 & 40.0 & 10.00 \\
7 & 4 & 60.0 & 34.3 & 8.57  \\
8 & 4 & 60.0 & 30.0 & 7.50  \\
\hline
4 & 8 & 30.0 & 60.0 & 7.50  \\
5 & 8 & 30.0 & 48.0 & 6.00  \\
6 & 8 & 30.0 & 40.0 & 5.00  \\
7 & 8 & 30.0 & 34.3 & 4.29  \\
8 & 8 & 30.0 & 30.0 & 3.75  \\
\hline
\end{tabular}
\end{table}

\begin{table}[t]
\centering
\caption{\textbf{Scene diversity scaling data composition.} Total training budget (minutes) for different scene counts and data-usage fractions (relative to 60 min/scene at 100\%). We evaluate all models on unseen demonstrators at unseen scenes.}
\label{tab:scene_diversity_budget}
\setlength{\tabcolsep}{4pt}
\renewcommand{\arraystretch}{1.12}
\footnotesize
\begin{tabular}{c|ccccc}
\hline
\multicolumn{1}{c|}{\textbf{\# Scenes}} & \multicolumn{5}{c}{\textbf{Data usage fraction}} \\
\multicolumn{1}{c|}{} & \textbf{6.25\%} & \textbf{12.5\%} & \textbf{25\%} & \textbf{50\%} & \textbf{100\%} \\
\hline
\textbf{1}  & 3.75 & 7.5  & 15  & 30  & 60  \\
\textbf{2}  & 7.5  & 15   & 30  & 60  & 120 \\
\textbf{4}  & 15   & 30   & 60  & 120 & 240 \\
\textbf{8}  & 30   & 60   & 120 & 240 & 480 \\
\textbf{16} & 60   & 120  & 240 & 480 & 960 \\
\hline
\end{tabular}

\vspace{3pt}
{\footnotesize \textit{All values are minutes of recording time for training data.}}
\end{table}

\subsection{Latent Space Visualization}
\label{appendix:L}
We choose UMAP over t-SNE for dimensionality reduction because, while both preserve local neighborhood structure, UMAP also retains a degree of global structure.

\textbf{EgoVerse-I Dataset Visualization} For the EgoVerse-I dataset visualization, we load all training data in sequential order and sample uniformly (selecting every 15 datapoints).The data points are then passed into DinoV3~\cite{siméoni2025dinov3} large model which returns image embeddings. These image embeddings then have their dimension reduced to 2 with UMAP, which we visualized in Fig.~\ref{fig:umap_diversity}.

\textbf{Demonstrator Diversity Visualization} For the demonstrator diversity visualization, we load all training data in sequential order and also sample uniformly (selecting every 15 datapoints). The datapoints are then passed into the trained HPT model. We then take the 64 action conditioned tokens that condition the Flow Matching Action Decoder and flatten them into a single latent vector. UMAP is then applied on these latent vectors to generate 2 dimensional vectors which we visualized in Fig.~\ref{fig:diversity_tsne}.

\label{appendix:N}

\end{document}